\documentclass[final]{cvpr}
\usepackage{times}
\usepackage{epsfig}
\usepackage{graphicx}
\usepackage{amsmath}
\usepackage{amssymb}

\usepackage{mathrsfs}
\usepackage[ruled,vlined]{algorithm2e}
\usepackage{color}
\usepackage{booktabs}
\usepackage{multirow}
\usepackage{mathtools, bm, bbm}
\usepackage{bbding}
\usepackage[table]{xcolor}
\usepackage{pifont}
\usepackage{comment}

\DeclareMathOperator*{\argmin}{arg\,min}

\definecolor{mygray}{gray}{0.45}
\definecolor{violet}{rgb}{0.54, 0.17, 0.89}
\definecolor{cadmiumorange}{rgb}{0.93, 0.53, 0.18}

\usepackage[pagebackref=true,breaklinks=true,colorlinks,bookmarks=false]{hyperref}

\begin{document}
\pagestyle{empty}
\title{RPSRNet: End-to-End Trainable Rigid Point Set Registration Network
using Barnes-Hut $\mathbf{2^D}$-Tree Representation}

\author{
Sk Aziz Ali$^{1,2}$ $\;\;\;\;\;$  
Kerem Kahraman$^{1}$  $\;\;\;\;\;$
Gerd Reis$^{2}$ $\;\;\;\;\;$
Didier Stricker$^{1,2}$\vspace{2pt}\\
$^{1}$TU Kaiserslautern$\;\;$
$^{2}$German Research Center for Artificial Intelligence (DFKI GmbH), Kaiserslautern$\;\;$
}

\maketitle
\thispagestyle{empty}

\begin{abstract}
We propose RPSRNet - a novel end-to-end trainable deep neural network for rigid point set registration. 
For this task, we use a novel $2^\text{D}$-tree representation for the input point sets and a hierarchical deep feature embedding in the neural network.
An iterative transformation refinement module of our network boosts the feature matching accuracy in the intermediate stages.
We achieve an inference speed of ${\sim}$12-15$\,$ms to register a pair of input point clouds as large as ${\sim}$250K. 
Extensive evaluations on (i) KITTI LiDAR-odometry and (ii) ModelNet-40 datasets show that our method outperforms prior state-of-the-art methods -- \eg, on the KITTI dataset, DCP-v2 by 1.3 and 1.5 times, and PointNetLK by 1.8 and 1.9 times better rotational and translational accuracy respectively.
Evaluation on ModelNet40 shows that RPSRNet is more robust than other benchmark methods when the samples contain a significant amount of noise and disturbance.
RPSRNet accurately registers point clouds with non-uniform sampling densities, \eg, LiDAR data, which cannot be processed by many existing deep-learning-based registration methods. 
\end{abstract}

\vspace{-0.25cm}
\section{Introduction}\label{sec:Introduction}
Rigid point set registration (RPSR) is indispensable in numerous computer vision and graphics applications -- \eg, camera pose estimation~\cite{aldoma2011cad, Umeyama1991}, LiDAR-based odometry~\cite{Zhang-2016-110808, li2019net}, 3D reconstruction of partial scenes~\cite{gross2019alignnet}, simultaneous localization and mapping tasks~\cite{KinextFusion_Newcombe}, to name a few. 
An RPSR method estimates the rigid motion field, parameterized by rotation ($\mathbf{R}\in \mathcal{SO}(D)$) and translation ($\mathbf{t} \in \mathbb{R}^D$), of a moving sensor from the given pair of $D$-dimensional point clouds (\hbox{\textit{source}}
and \hbox{\textit{target}}). 

\begin{figure}[ht]
\begin{center}
\includegraphics[width=3.35in]{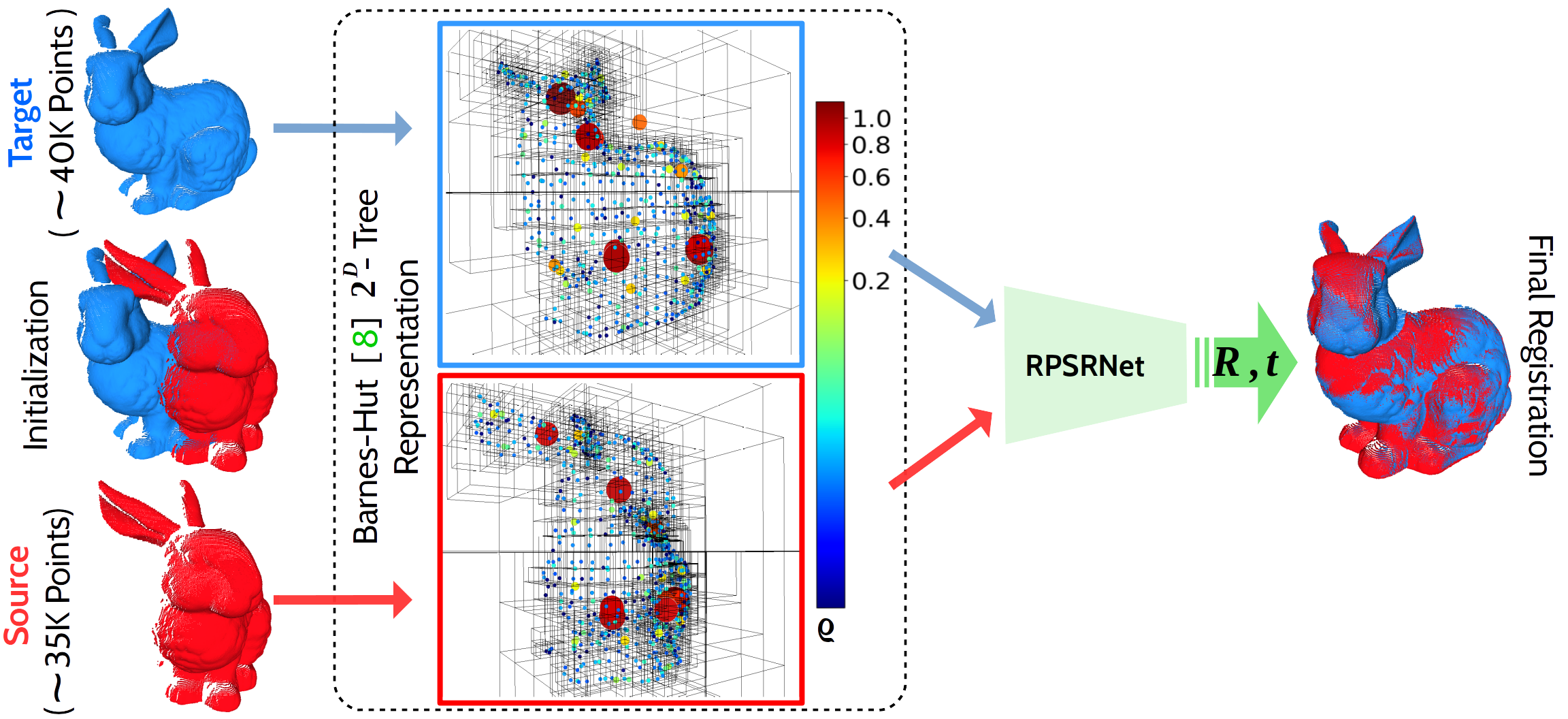}
\end{center}
\vspace{-0.25cm}
\caption{\textbf{Rigid Point Set Registration using Barnes-Hut (BH) $\mathbf{2^D}$-tree Representation.} The center-of-masses (CoMs) and point-densities ($\varrho$) of non-empty tree-nodes are computed for the respective BH-trees of the \textit{source} and \textit{target}. These two attributes are input to our RPSRNet which has global feature-embedding blocks employing tree-convolution operations on the nodes. Finally, we regress rigid rotation ($\mathbf{R}\in \mathcal{SO}(D)$) and translation ($\mathbf{t}\in\mathbb{R}^D$) parameters from output features.
}
\label{fig:TEASER_RPSRNet}
\end{figure}
Generally, different types of input data from diverse application 
areas pose distinct challenges to registration methods, \eg, (i) 
LiDAR data contains large number of points with non-uniform sampling density, (ii) partial scans obtained from structured-light sensors or multi-view camera systems contain a large number of points with small amount of overlap between the point clouds, (iii) RGB-D sensors, such as Kinect, yield dense depth data with large displacement between consecutive frames.

\vspace{0.05cm}
\noindent\textbf{Classical RPSR Methods. }
Iterative Closest Point  (ICP) \cite{BeslMcKay1992} and its many variants~\cite{Greenspan2003,Rusinkiewicz2001,Korn2014ColorSG,GeneralizedICPSegal,Yang2013, EM-ICP_Granger, GeomStableSamplingICP2003} are the most widely used methods that alternate the nearest correspondence search and transformation estimation steps at every iteration.
A comparative study~\cite{Pomerleau2013} shows that several ICP variants target a specific challenge, and many of them are prone to converge in bad local-minima.
Coherent Point Drift (CPD)~\cite{MyronenkoSong2010} 
is another state-of-the-art approach from the class of probabilistic RPSR methods.
Similar to CPD, GMMReg~\cite{GMMReg1544863}, FilterReg~\cite{Gao2019}, and
HGMR~\cite{Eckart2018HGMRHG} are also probabilistic RPSR methods that treat the data as Gaussian mixture models.
The probabilistic models for rigid alignment gives more robustness against noisy input than ICP~\cite{BeslMcKay1992}.
Recently physics-based approaches, \eg, GA~\cite{Golyanik2016}, BH-RGA~\cite{BHRGA2019}, FGA~\cite{2020arXiv200914005A},
and~\cite{Agarwal20173DPC, Ali2018NRGAGA}, are appearing computationally faster and more robust than ICP or CPD. 
These methods assume that the point clouds are astrophysical particles with masses and obtain the optimal alignment by defining a motion model for the source in a simulated
gravitational field. 
However, most of the above methods run on CPUs and do not scale to register large point clouds, \eg, LiDAR scans.
Due to this, often handcrafted~\cite{BSC2017} or automatically extracted \cite{Rusu2008, Rusu2009} feature descriptors are iteratively sampled using RANSAC~\cite{raguram2008comparative} to obtain true correspondence matches. On the other hand, Fast Global Registration (FGR)~\cite{FGReccv16} replaces the RANSAC step with a robust global optimization technique to obtain true matches.

\vspace{0.05cm}
\noindent\textbf{DNN-based Point Processing Registration Methods. }
A recent survey~\cite{s19194188} shows that many contributions are
made to the development of deep neural networks (DNNs) for point cloud processing (PCP) tasks~\cite{qi2017pointnet,qi2017pointnet++,wu2019pointconv,AokiGSL19,wang2019deep,wang2019prnet,lu2019deepvcp,gross2019alignnet,Huang_2020_CVPR,choy2020deep,tatarchenko2017octree,RPMNet2020,3DRegNetCVPR19,Nezhadarya_2020_CVPR, liu2019densepoint,su18splatnet,liu2019rscnn}, \eg, classification and segmentation~\cite{qi2017pointnet, qi2017pointnet++, wu2019pointconv}, correspondence and geometric feature matching~\cite{gojcic2019perfect, Khoury2017LearningCG}, up-sampling~\cite{yifan2019patch}, and down-sampling~\cite{Nezhadarya_2020_CVPR}.
Rigid point-set registration using DNNs appeared recently~\cite{Elbaz20173DPC,AokiGSL19,wang2019deep,wang2019prnet,lu2019deepvcp,gross2019alignnet,3DRegNetCVPR19,RPMNet2020,choy2020deep}.
Elbaz~\etal~\cite{Elbaz20173DPC} introduce the LORAX algorithm for large scale point set registration using super-point representation.
PointNetLK~\cite{AokiGSL19} is the first RPSR method which uses PointNet~\cite{qi2017pointnet} (w/o its T-net component) to obtain the global
feature-embedding of \textit{source} and \textit{target}. The method further uses the iterative Lucas and Kanade (LK) method to obtain the transformation optimizing the distance between global features.
Unlike PointNetLK, the recent RPSR method Deep Closet Point (DCP)~\cite{wang2019deep} chooses DGCNN~\cite{wang2019dynamic} to obtain the feature embedding and a Transformer network~\cite{devlin2018bert} to learn contextual residuals between them.
%
%
%
PRNet~\cite{wang2019prnet} is an extension of DCP~\cite{wang2019deep} 
to solve a matching sharpness issue. 
%
%
%
%
Deep Global Registration (DGR)~\cite{choy2020deep}, \cite{DengCVPR2019}, and  3DRegNet~\cite{3DRegNetCVPR19} are among the latest methods to report better accuracy than FGR~\cite{FGReccv16} for registering partial-to-partial scans. DGR takes fully convolutional geometric features (FCGF)~\cite{choy2019fully} and trains a high dimensional convolutional network~\cite{choy2020high} to classify inliers/outliers from the input FCGF descriptors. Finally, a weighted Procrustes~\cite{gower1975generalized} using inlier weights estimate the transformation. 3DRegNet~\cite{3DRegNetCVPR19} method also employs a classifier, similar to the inlier/outlier classification block of DGR, but using deep ResNet~\cite{DeepResNet2016} layers, followed by differentiable Procrustes~\cite{gower1975generalized} to align the scans. Notably, both 3DRegNet and DGR have ${\sim}4$ times and ${\sim}10$ higher runtime than DCP~\cite{wang2019deep}. Some other RPSR methods using DNNs are AlignNet-3D~\cite{gross2019alignnet}, DeepGMR~\cite{yuan2020deepgmr} and DeepVCP~\cite{lu2019deepvcp}.

\vspace{0.05cm}
\noindent\textbf{Problems in DNN-based Point Cloud Processing. }
Generally, DNNs on 3D point clouds give more intuitive high  dimensional geometric features learned from the given samples. 
Although, the convolution operations are not straight forward for DNNs on point clouds 
because they can be \textit{unordered, irregular} and \textit{unstructured}. 
Voxel-based~\cite{maturana2015voxnet} or shallow grid-based~\cite{riegler2017octnet} representations use volumetric convolution which is very memory demanding ($\mathcal{O}(N^3)$ where $N$ is the voxel resolution) and can only be applied to very small problems. 
In contrast, multilayer perceptron (MLP) based convolution in~\cite{qi2017pointnet, qi2017pointnet++, wu2019pointconv,yifan2019patch} operate on sub-sampled versions of the point clouds. Thus, the local correlation between the latent-features are 
inefficiently established using random neighborhood search in MLP-based methods. Another critical disadvantage of methods relying on PointNet~\cite{qi2017pointnet} (or its extension~\cite{qi2017pointnet++}) is that deconvolution is inapplicable. RPMNet~\cite{RPMNet2020}, which is another PointNet~\cite{qi2017pointnet} reliant RPSR method, shows that it requires points' normals to be computed beforehand for robust alignment. Additionally, the inference time, a crucial parameter for real-time applications, of several recent DNN-based methods~\cite{RPMNet2020, 3DRegNetCVPR19, choy2020deep} are not on a par with DCP~\cite{wang2019deep} or DeepVCP~\cite{lu2019deepvcp}.     
Among the DNN based approaches of RPSR, no method is available so far which addresses the aforementioned issues.

\vspace{0.05cm}
\noindent\textbf{$\mathbf{2^D}$-Tree Based Methods for Point Cloud Processing.}
$2^\text{D}$-tree (\eg, octree in 3D or quadtree in 2D) representation~\cite{zhou2010data} of a point cloud is more memory efficient to encode local correlation among the points inside a tree-cell as they can share the same input signals. 
Unlike regular grids, $2^\text{D}$-tree representation free up the empty cells, which are often more than $50\%$ for sparse point clouds, from being computed. 
Moreover, it provides hierarchical correlations between the neighborhood cells. 
FGA~\cite{2020arXiv200914005A} and BH-RGA~\cite{BHRGA2019} are the latest physics-based RPSR methods to use Barnes-Hut (BH) $2^\text{D}$-tree representation of point clouds. 
FGA shows state-of-the-art alignment accuracy and the fastest speed among the classical methods. 
DNN-based methods for shape reconstruction from a single image~\cite{tatarchenko2017octree}, point cloud classification and segmentation (PCCS)~\cite{lei2019octree, Wang-2017-ocnn}, real-time 3D scene analysis~\cite{OctreeNet2020}, shape retrieval~\cite{Riegler2017CVPR}, and large LiDAR data compression~\cite{Huang_2020_CVPR} have all appeared in last three years which claim $2^\text{D}$-tree as more efficient learning representation for point clouds. 
Although, there is no RPSR method that addresses the aforementioned problems of previous learning-based approaches and simultaneously utilizes the efficacy of $2^\text{D}$-tree representation. 
%
%

In this paper, we present the first DNN-based RPSR method using a novel Barnes-Hut (BH)~\cite{1986Natur324446B} $2^\text{D}$-tree representation of input point clouds (see Fig.~\ref{fig:TEASER_RPSRNet}). 
At first, we build a BH-tree by recursively subdividing the normalized bounding space of an input point cloud up to a limiting depth $d$ (Sec.~\ref{sec:ProposedBHtreeMethod}).   
A tree with maximum depth $d = 6$ or 7 (\ie, equivalent to $64^3$ or $128^3$ voxels in 3D) gives a fine level of granularity for our proposed DNN to process. %
Except from the root node, which contains all the points, the internal nodes of the $2^{\text{D}}$-tree encapsulate varying number of points.
Hence, the center-of-masses (CoMs) and the inverse densities (IDs) of the nodes, computed at each depth, are the representative attributes of a BH-tree. 
Besides, the neighbors of a given node  are easily retrievable using established indexing and hash maps~\cite{bader2012space} for $2^\text{D}$-tree. 
With this input representation, Sec.~\ref{sec:RPSRNet_Architecture_Detail} describes the complete pipeline of RPSRNet. 
To this end, we design a single DNN block -- \textit{hierarchical feature extraction} (HFE) -- with $d$ hidden layers for global feature extraction.
Two sub-blocks under HFE -- namely \textit{hierarchical position feature embedding} (HPFE) and \textit{hierarchical density feature embedding} (HDFE) respectively -- learn the positional and density features, respectively. 
We apply late-fusion between HPFE and HDFE that results in a density adaptive embedding. 
As a result, learned-features become homogeneous for the input point clouds with non-uniform point sampling densities (\eg, LiDAR scans).   
The final block of our RPSRNet, inspired by DCP-v2~\cite{wang2019deep}, contains a relational network~\cite{santoro2017simple}\footnote{to extract the correspondence weights from our hierarchical deep feature embedding of source and target} with an integrated transformer~\cite{Transformer2017}, and a differentiable singular value decomposition (SVD)\footnote{to estimate rotation and translation parameters.} module.
We further refine the estimation with an iterative rigid alignment architecture with multiple alignment-passes for a single pair of input sample (See Fig.~\ref{fig:fig_RPSRNetArchiteture} and Sec.~\ref{subsec:IterativeRegNet}). 
%
%

%
%

\vspace{0.15cm}
\noindent\textbf{Contributions.} The overall contributions and promising characteristics of this work are as follows:
\vspace{-0.2cm}
\begin{itemize}
 \item A novel BH $2^\text{D}$-tree representation of input point sets. 
 \vspace{-0.25cm}
 \item An end-to-end trainable rigid registration network with real time inference on dense point clouds using the following components: 
    \vspace{-0.15cm}
    \begin{enumerate}
        \item HPFE and HDFE using $2^\text{D}$-tree convolution
        \vspace{-0.15cm}
        \item Rigid transformation estimation using relational network and differentiable SVD. 
        \vspace{-0.15cm}
        \item A multi-pass architecture for iterative transformation ($\mathbf{R}, \mathbf{t}$) refinement (similar to~\cite{wang2019prnet})
    \end{enumerate}
\end{itemize}

\section{The Proposed BH $\mathbf{2^D}$-Tree Construction}\label{sec:ProposedBHtreeMethod}
\begin{figure}[!ht]
\begin{center}
\includegraphics[width=0.99\linewidth]{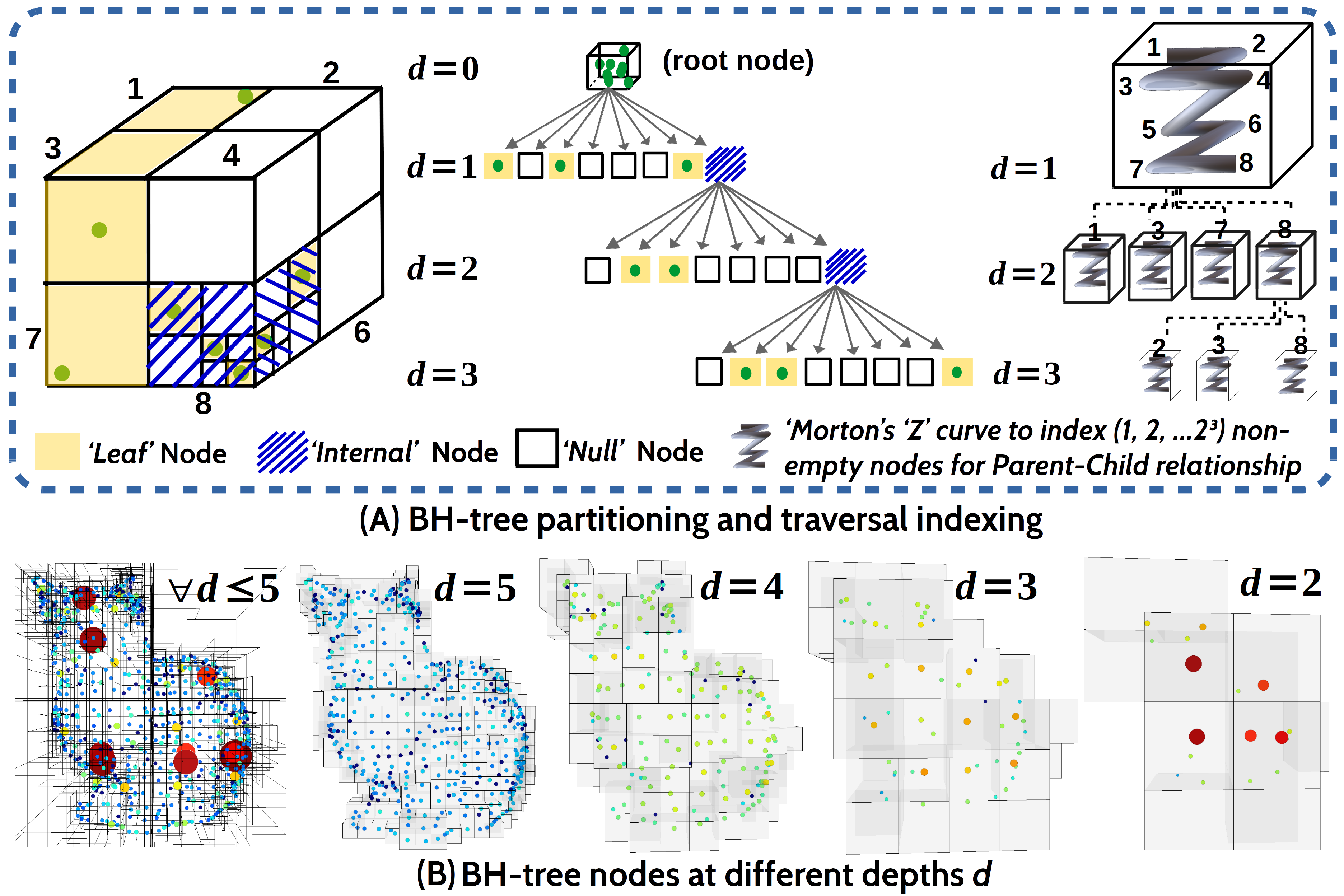} 
\end{center}
\vspace{-0.25cm}
\caption{Barnes-Hut $\mathbf{2^D}$-Tree construction and nodes.}
\label{fig:Barnes-HutTreeVisulization}
\end{figure}
Given a pair of misaligned input point clouds, a source 
$\mathbf{Y}\in\mathbb{R}^{M\times\text{(D+c)}}$ 
and a target 
$\mathbf{X}\in\mathbb{R}^{N\times\text{(D+c)}}$, 
we build Barnes-Hut (BH)~\cite{1986Natur324446B} $\mathbf{2^D}$-trees 
$\mathbf{\tau^{\mathbf{Y}}}$ and $\mathbf{\tau^{\mathbf{X}}}$ 
on them. 
The trees
($\mathbf{\tau^{\mathbf{Y}}}$ and $\mathbf{\tau^{\mathbf{X}}}$) are defined by their respective sets of nodes 
($\mathbf{N}_{d}^{\mathbf{Y}}$ and $\mathbf{N}_{d}^{\mathbf{X}}$) and the parent-child relationships among them at different depths $d = 1, 2,\hdots$. 
Every point in $\mathbf{Y}$ and $\mathbf{X}$ is a D-dimensional position vector, often associated with an additional c-dimensional input feature channel.
The input features can be RGB color channels, local point densities, or other attributes. 
Although BH-tree is generalizable for D-dimensional input data, we describe further details and notations \hbox{w.r.t} 3D data (\ie $\text{D}=3$) for easier understanding. 
The following steps describe how to build a BH octree (\ie $2^{\text{3}} = 8$ child octants per parent node) by recursively subdividing the Euclidean space of an input:

\vspace{-0.2cm}
\begin{enumerate}
 \item \label{BH:Step1} We determine the extreme point positions, \ie, $\min$, $\max$ values along x, y, and z-axes, and split the whole space (as root node) from its center into $2^3$ cubical sub-cells as its child nodes. The \textit{half-length} of the cube is called cell length $r$. Depending on the structure of the input point clouds, if not a regular grid structure, the child nodes can be \textit{empty}, \textit{non-empty with more than one point}, and \textit{non-empty with exactly one point}. We call them \textit{null}, \textit{internal}, and \textit{leaf} nodes (see Fig.~\ref{fig:Barnes-HutTreeVisulization}-A).

 \vspace{-0.2cm}
 \item \label{BH:Step2} For each \textit{internal} node, we compute the center-of-mass (CoM) $\mu\in\mathbb{R}^3$ and inverse density (ID)\footnote{inspired by Monte-Carlo Importance Sampling Density -- if $m$ random samples drawn from a meta-distribution function $q(x)$ of a distribution $p(x)$, then expectation of sample can be expressed as a fraction of weights $w(x) = \frac{p(x)}{q(x)}$, such that the normalized importance is $\sum_{i=1}^m w(x_i) = 1$.} $\varrho^-\in\mathbb{R}^+$ of the encapsulated points inside. These two attributes (CoM and ID) of a \textit{leaf} node are equivalent to the point position and its local density. 
 \vspace{-0.15cm} 
 \item \label{BH:Step3} We continue the subdivision of the \textit{internal} nodes only up to a maximum depth $d_{0}$ if \textit{leaf} node is not reached.
\end{enumerate}
Fig.\ref{fig:Barnes-HutTreeVisulization}-(B) illustrates the BH-tree CoMs of the nodes at different depths $d=2, 3, 4$ and $5$ with their color-coded (magnitude order: red $>$ green $>$ blue) point densities $\varrho$.
Furthermore, we establish the indexing of the nodes ($\mathbf{N}_{d}$) as well as 
their neighborhood indices ($\kappa_d$) at every depth $d$ using a label array ($L_d$) as hash key 
to retrieve the parent-child information. For instance, the child nodes at depth $d$ of 
$\mathbf{N}_{d-1,i}$ (a non-empty parent node $i$ at depth $d-1$) will be indexed in order 
of the Morton's 'Z' curve: $\{2^3\cdot i + 1,\hdots, 2^3\cdot i + 8\}$. 
The supplement has exemplary indexing.  

\vspace{-0.2cm}
\section{The Proposed RPSRNet Method}\label{sec:RPSRNet_Architecture_Detail}
In this section, we formulate the rigid alignment problem between two BH-trees constructed on the input point cloud pair, and then describe our RPSRNet model for this rigid alignment task by highlighting its different components.  
\begin{figure}[ht]
\begin{center}
\includegraphics[width=0.99\linewidth]{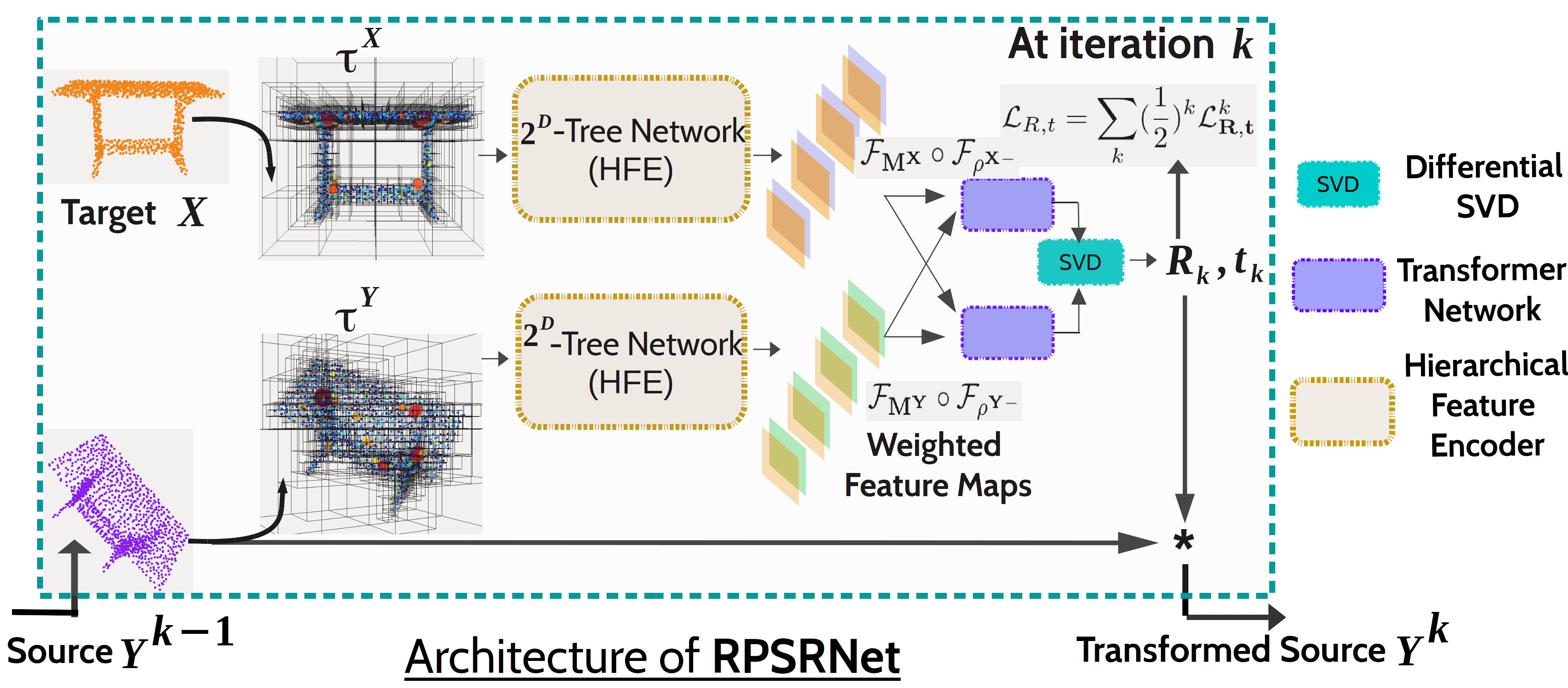}
\end{center}
\vspace{-0.25cm}
\caption{A schematic design of iterative RPSR network using $2^{\text{D}}$ tree representation of input point clouds.}
\label{fig:fig_RPSRNetArchiteture}
\end{figure}

\vspace{-0.15cm}
\subsection{Mathematical Formulation}
\label{subsec:MathFormulation}
%
%
Rigid registration of a movable 3D point cloud $\mathbf{Y} = \left\lbrace  \mathbf{y}_1,\hdots,\mathbf{y}_M \right\rbrace \in\mathbb{R}^{M\times3}$ (\ie, \textit{source}) with another point cloud $\mathbf{X}= \left\lbrace  \mathbf{x}_1,\hdots,\mathbf{x}_N \right\rbrace \in\mathbb{R}^{N\times3}$ (\ie, \textit{target}) is to estimate a 6DoF transformation $\mathbf{T} = \left[\mathbf{R}\in\mathcal{SO}(3)|\mathbf{t}\in\mathbb{R}^3\right]$, which can bring both $\mathbf{Y}$ and $\mathbf{X}$ in a common reference frame. 
The problem is often formulated as optimization of cost function $\mathbf{U(R,t)}$ in the form of globally multiply-linked correspondence distance errors: 
\begin{equation}
\vspace{-0.2cm}
\small
\mathbf{U(R,t,X,Y)} = 
\sum_{i,j} \mathbf{\omega}_{ij}\lVert(\mathbf{R} \mathbf{y}_i + \mathbf{t}) - \mathbf{x}_j\rVert_{2}^{2}, 
\label{eqn:DistanceTransform}
\end{equation}
where $\lVert\cdot\rVert$ denotes the $\ell_2$-norm of the distance, and $\mathbf{\omega}_{ij}$ is correspondence weight. Without $\mathbf{\omega}_{ij}$ (or with constant values for all ($\mathbf{y}_i,\,\mathbf{x}_j$) pairs), the influence of putative matches on Eq.~\eqref{eqn:DistanceTransform} cannot be distinguished from spurious matches. 
In the context of our BH-tree representation of the inputs (\ie, 
$\mathbf{Y} \rightarrow \mathbf{\tau^{\mathbf{Y}}}$ and 
$\mathbf{X} \rightarrow \mathbf{\tau^{\mathbf{X}}}$),
we reformulate $\mathbf{U(R,t)}$ as a multi-scale sum of mean-squared distance errors between the CoMs ($\mathbf{M}_{d}^{\mathbf{Y}} = \{\mathbf{\mu}_{d,l}^{y}\}$ and $\mathbf{M}_{d}^{\mathbf{X}} = \{ \mathbf{\mu}_{d,l}^{x}\}$) of the respective sets of non-empty tree-nodes ($\mathbf{N}_{d}^{\mathbf{Y}}=\{\mathbf{n}_{d,l}^{y}\}$ and 
$\mathbf{N}_{d}^{\mathbf{X}}=\{\mathbf{n}_{d,l}^{x}$\}) at different depths $d\in\{1,2,\hdots,d_0\}$ and labels $l\in\{1,2,..,(2^3)^d\}$), and weighted by their corresponding ID values $\mathbf{\rho}_{d}^{\mathbf{Y-}}=\{\varrho_{d,l}^{y-}\}$ and 
$\mathbf{\rho}_{d}^{\mathbf{X}-}=\{\varrho_{d,l}^{x-}\})$. Therefore, the multiply-linked ($\sum_{i,j}$) correspondence distance errors are now applicable on the CoMs of the non-empty nodes at every depth:
\begin{equation}
\vspace{-0.2cm}
\footnotesize
\mathbf{U(R,t,\tau^{\mathbf{X}},\tau^{\mathbf{Y}})} = 
    \sum_{d}\sum_{l,\hat{l}}\varrho_{d,l}^{y-}\varrho_{d,\hat{l}}^{x-}\lVert \left(\mathbf{R}\mu_{d,l}^{y} + t\right) - \mu_{d,\hat{l}}^{x} \rVert_{2}^{2}.
\label{eq:TreeDistanceTransform}
\end{equation}
Generalized Procrustes methods~\cite{Kabsch1976, gower1975generalized} give a closed-form solution for $\hat{\mathbf{R}}, \hat{\mathbf{t}} = \argmin_{\mathbf{R,t}} \mathbf{U(R,t,X,Y)}$ in Eq.~\eqref{eqn:DistanceTransform} without input weights $\omega_{ij}$ as $\hat{\mathbf{R}} = \mathbf{USV^{\text{T}}}$ and $\hat{\mathbf{t}} = \mathbf{X} - \mathbf{RY}$, where $\mathbf{U\Sigma  V^{\text{T}}} = \text{SVD}((\mathbf{X}-\overline{\mathbf{X}})(\mathbf{Y}-\overline{\mathbf{Y}})^{\text{T}})$ and $\mathbf{S} = \textit{diag}(1,\hdots,1,\text{det}(\mathbf{U}\mathbf{V}))$. $\overline{\mathbf{X}}$ and  $\overline{\mathbf{Y}}$ are the mean of the inputs $\,\,\square$. Analogously, the DGR~\cite{choy2020deep} method shows a closed form solution for $\mathbf{R,t}$ with correspondence weights $\omega_{ij}$ in Eq.~\eqref{eqn:DistanceTransform} 

%
%
Our RPSRNet is a deep-learning framework (see Fig.~\ref{fig:fig_RPSRNetArchiteture}) to estimate rigid transformation $\mathbf{R,t}$ in Eq.~\eqref{eq:TreeDistanceTransform} by using high-dimensional features learned for the CoM and ID attributes of input trees. Our pipeline has three main stages:

\vspace{0.05cm}
\noindent\textbf{Stage 1:}
We propose a hierarchical feature encoder (HFE) that returns feature maps $\mathcal{F}_{\tau^{\mathbf{Y}}}$ and $\mathcal{F}_{\tau^{\mathbf{X}}}$ for the input trees. 
The output feature maps from HFE are obtained by fusing (Hadamard product) positional features with features from inverse density channel. Note, that the fused feature maps (position $\circ$ density)  $\mathcal{F}_{\mathbf{M^{\mathbf{Y}}}}\circ\mathcal{F}_{\rho^{\mathbf{Y-}}}$ and $\mathcal{F}_{\mathbf{M^{\mathbf{X}}}}\circ\mathcal{F}_{\rho^{\mathbf{X-}}}$ are computed separately. 

\vspace{0.1cm}
\noindent\textbf{Stage 2:} 
In this stage, we use a Transformer network~\cite{Transformer2017} to learn the contextual residuals $\phi(\mathcal{F}_{\tau^{\mathbf{Y}}}, \mathcal{F}_{\tau^{\mathbf{X}}})$ and $\phi(\mathcal{F}_{\tau^{\mathbf{X}}}, \mathcal{F}_{\tau^{\mathbf{Y}}})$, between the two embedding $\mathcal{F}_{\tau^{\mathbf{Y}}}$ and $\mathcal{F}_{\tau^{\mathbf{X}}}$. 
The contextual map $\phi:\mathbb{R}^{64\times512} \times \mathbb{R}^{64\times512} \rightarrow \mathbb{R}^{64\times512}$ denotes the changes between two input embedding (see dimension of output feature from HFE in Sec.~\ref{subsec:HFE_block}). 

\vspace{0.1cm}
\noindent\textbf{Stage 3:} We use the differential SVD~\cite{AutoDiffPytorch} block on the final score matrix as suggested by DCP~\cite{wang2019deep}: 
\begin{equation}
    \small
    S = \text{SoftMax}(\left[\mathcal{F}_{\tau^{\mathbf{X}}} + 
    \phi(\mathcal{F}_{\tau^{\mathbf{X}}}, \mathcal{F}_{\tau^{\mathbf{Y}}})\right]
    \left[\mathcal{F}_{\tau^{\mathbf{Y}}} + 
    \phi(\mathcal{F}_{\tau^{\mathbf{Y}}}, \mathcal{F}_{\tau^{\mathbf{X}}})\right]^\text{T}
    )
\end{equation}

\vspace{-0.3cm}
\subsection{Hierarchical Feature Encoder}\label{subsec:HFE_block}
\vspace{-0.25cm}
\begin{figure}[ht]
\begin{center}
\includegraphics[width=0.99\linewidth]{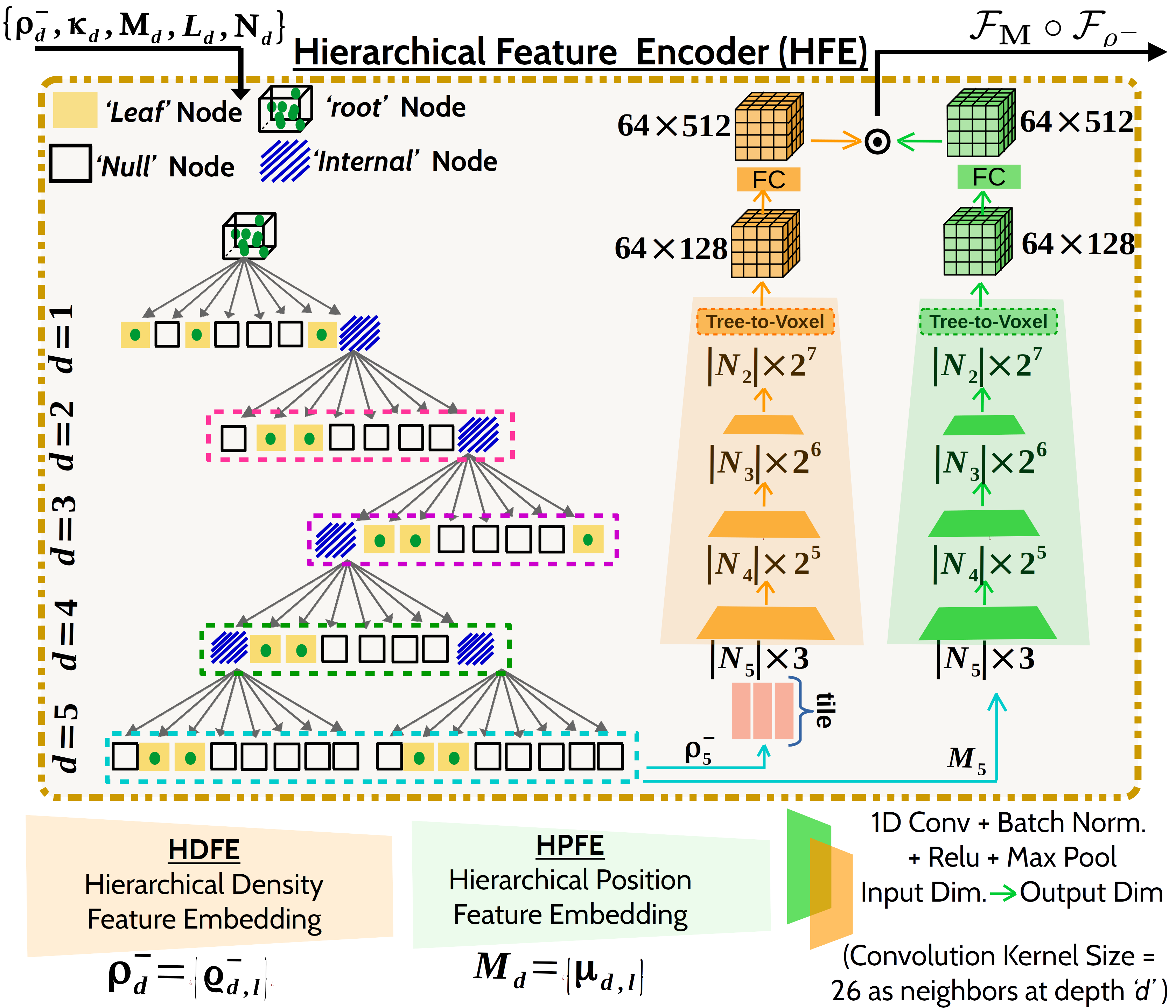}
\end{center}
\vspace{-0.25cm}
\caption{Nodes at different depth of Barnes-Hut tree as the input of our hierarchical feature encoder (HFE).
}
\label{fig:fig_HFEArchiteture}
\end{figure}
\vspace{-0.1cm}
%
%
%
Our HFE (see Fig.~\ref{fig:fig_HFEArchiteture}) takes the BH $2^3$-tree attributes -- the CoMs, IDs, 
labels of nodes, and list of 26 neighbours of all non-empty nodes at every depth as input. HFE has has two embedding layers -- one to encode the features 
from the position of CoMs, and the other from the IDs -- named as HPFE and HDFE, respectively. 
%
%
%
The encoder has four hidden layers. \textit{Three} of them encode the features only at 
the non-empty ($\mathbf{N}_{d}^{\mathbf{Y}}$, $\mathbf{N}_{d}^{\mathbf{Y}}$) at 
depths $d=5, 4, 3,$, and the \textit{fourth} is \textit{Tree-to-Voxel} layer which converts 
the non-empty tree nodes at depth 2 into a voxel structure, \ie, $2^3\cdot 2^3 = 4^3$. The 
empty locations are zero-filled. The final layer is a \textit{fully connected layer} to 
up-sample the feature maps.
Each embedding layer does a set of operations, \ie, 1D convolution + 
Batch Normalization~\cite{BatchNorm2015} + ReLU~\cite{ReLuVinod2010} + 
Max-Pooling operations. We call this as a basic unit and named HDFE($d$) or HPFE($d$) 
based on their input attributes.

\noindent\textbf{Barnes-Hut $\mathbf{2^3}$-Tree Convolution / Pooling.}\label{subsec:TreeConv}
To perform the tree convolution operation, we use smart indexing of tree nodes, 
parent-child relation order, and adjacency system of their neighborhoods. 
Fig.~\ref{fig:Barnes-HutTreeVisulization}-(A) shows a \lq Z\rq
-curve (known as Morton's curve) traverse through nodes at the current depth 
$d$ and increase their label index $l$ by 1 leaving the empty node label as -1. 
If the parent node was empty, the \lq Z\rq  curve skips that octant. 
Hence, we keep an array for the 
labels $L_d$ at every depth to serve as hash map (See supplementary material ). Next, for the convolution operation on every 
\textit{non-empty} node, 26 adjacent nodes from the same depth are fetched. 
Notably adjacent sibling nodes can be empty. At run time, we dispatch a 
\textit{zero filling} operation on those empty neighbors for indexed based 
($L_d$) 1D convolution. In the reverse case, while Max-Pooling, information 
flows from child to parent nodes. Hence the same operation of zero filling is 
done on the empty siblings of a non-empty parent 
(See supplement for for detailed example indexing on a tree and operations 
like zero-filling and child-filling). The following is the input and output 
feature dimensions at depth $d$ using weight filter $W$ after 1D convolution: 
\begin{equation}\label{eq:embeddingBlock}
\small
 \mathcal{F}_{\textit{out}} = \text{Conv1D}
  (
   W\in\mathbb{R}^{|\mathbf{N_d}|\times 26+1} , 
   \mathcal{F}_{\textit{in}} \in \mathbb{R}^{|\mathbf{N_d}|\times 2^{(10-d)}}
  )
\end{equation}

\subsection{Iterative Registration Loss}\label{subsec:IterativeRegNet}
RPSRNet predicts the final transformation in multiple steps. 
On every internal iteration $k$, a combined loss $\mathcal{L}_{\mathbf{R,t}}^k$ on rotation:
$\mathcal{L}_{\mathbf{R}}^k = \lVert
			 (\mathbf{R}_{\text{pred}}^{k})^\text{T}\mathbf{R}_{\text{gt}}
			      -
			      \mathbf{I}
			      \rVert^2$, 
and translation:
$
\mathcal{L}_{\mathbf{t}}^k =  
			      \lVert
			      (\mathbf{t}_{\text{pred}}^{k}) 
			      -
			      \mathbf{t}_{\text{gt}}
			      \rVert^2
$ is minimized. Our combined loss 
$\mathcal{L}_{\mathbf{R,t}}^k$ and total loss $\mathcal{L}_{R,t}$ using 
learnable scale parameters $\sigma_{\mathbf{R}}$ and $\sigma_{\mathbf{t}}$ 
(for balanced learning of the components) are:
\begin{gather}\label{Eq:Loss_Function}
\vspace{-0.2cm}
 \mathcal{L}_{\mathbf{R,t}}^k =  \text{exp}(-\sigma_{\mathbf{R}})
				 \mathcal{L}_{\mathbf{R}}^k + \sigma_{\mathbf{R}}
				 +
				 \text{exp}(-\sigma_{\mathbf{t}})
				 \mathcal{L}_{\mathbf{t}}^k + \sigma_{\mathbf{t}}
				 \\
\text{and}\,\mathcal{L}_{R,t} = \sum\limits_{k}(\frac{1}{2})^k\mathcal{L}_{\mathbf{R,t}}^k,
\, \text{where}\,k_{0} = \text{max iteration}\notag.
\end{gather}The scale parameters $\sigma_{\mathbf{R}}$ and $\sigma_{\mathbf{t}}$ help the network to
learn the wide range of transformations. 

\vspace{-0.1cm}
\section{Data Preparation and Evaluation Method}\label{sec:DataPrepEvalMethod}
\vspace{-0.15cm}
We evaluate on synthetic ModelNet40~\cite{ModelNet40} dataset which 
contains $9843$ training and $2468$ test samples of CAD models under 40 different categories, 
and also KITTI LiDAR-Odometry~\cite{Geiger2013IJRR, behley2019iccv} as 
several driving sequences.

\vspace{0.05cm}
\noindent\textbf{ModelNet40 Dataset.} 
In one setup \textbf{(M1)}, we use all the ${\sim}$9.8K training and ${\sim}$2.4K 
testing samples under all 40 different object categories. 
In another setup \textbf{(M2)}, to evaluate the generalizability of our network, 
we choose a training set $\mathcal{T}_{M}$ with shapes belonging to the first 
twenty categories, and testing set $\mathcal{T}_{M}^*$ where shapes are from the other 
twenty categories that are not seen during training.
%
%
%
All input samples are scaled between $[-1,1]$. 
The target point clouds are obtained after transforming its clone by random orientations ($\theta_x, \theta_y, 
\theta_z$) $\in\,(0^\circ, 45^\circ]$ and a random linear translation 
($\mathbf{t}_x, \mathbf{t}_y, \mathbf{t}_z$) $\in\,[-0.5, 0.5]$. 
To help the deep networks better cope with the data disturbances, another 950 
training and 240 testing samples 
are randomly selected to be preprocessed by four different settings 
of noise or data disturbances on source point clouds -- (i) adding Gaussian 
${\sim}\mathcal{N}(0, 0.02)$ and (ii) uniformly distributed noise 
${\sim}\mathcal{U}(-1.0, 1.0)$ which are $20\%$ of the total points in a sample, 
(iii) cropping a chunk (approx. $20\%$) of data, and (iv) jitter each point's 
position with a displacement tolerance 0.03. 
The choice of applying one of the four options is random.
We prepare five instances of the validation sets with increasing level of 
noise ($1\%, 5\%, 10\%, 20\%, \text{and} \,40\%$), jitter (increasing displacement 
threshold), and crops ($1\%, 10\%, 20\%, 30\%, \text{and} \,40\%$).

\vspace{0.05cm}
\noindent\textbf{KITTI Dataset.} 
%
%
There are 22 driving sequences in KITTI LiDAR odometry dataset. 
We prepare two setups -- in the first setup \textbf{(K1-w/o)}, the ground points 
are removed from each sample using the label information from SemanticKITTI
\cite{behley2019iccv}, and in the second setup \textcolor{mygray}{\textbf{(K2-w)}} 
samples remain unchanged.
The samples from each driving sequence 00 to 07 are split into $70\%,\, 
20\%,$ and $10\%$ as training, testing and validation sets and then merged. 
The number of frames in the sequences are $4541, 1101, 4661, 801, 271, 2761, 1101,\,$ 
and $1101$ respectively. 
Source point clouds from any of these sets are randomly selected frame-indices, whereas 
the corresponding targets are with next fifth frame-indices. 
Our RPSRNet can process point clouds with actual size. 
Due to the memory and scalability issues, PointNetLK~\cite{AokiGSL19}, 
DCP-v2~\cite{wang2019deep}, and CPD~\cite{MyronenkoSong2010} use inputs 
down-sampled to 2048 points.
%
%

%
%
\vspace{0.05cm}
\noindent\textbf{Evaluation Baselines.} 
We compare state-of-the-art neural network-based methods -- DCP-v2~\cite{wang2019deep}
and PointNetLK~\cite{AokiGSL19} against our RPSRNet. 
We also evaluate several unsupervised methods -- ICP~\cite{BeslMcKay1992}, FilterReg~\cite{Gao2019}, CPD~\cite{MyronenkoSong2010}, 
FGR~\cite{FGReccv16}, $\text{GA}\footnote[1]{\text{A GPU implementation}}$ \cite{Golyanik2016,BHRGA2019} -- for broader analysis (see Sec.~A of the supplement for details on training and parameter settings) and ignore some recent CNN-based methods~\cite{3DRegNetCVPR19, RPMNet2020, choy2020deep} that have higher run time ($>$ 150 milliseconds).  
%
%

%
%
\vspace{0.05cm}
\noindent\textbf{Evaluation Metrics.} We use angular deviation $\mathbf{\varphi}$ between the ground truth and predicted rotation matrices ($\mathbf{R}_{gt}, \mathbf{R}$), and similarly, the Euclidean distance 
error $\mathbf{\Delta t}$ between the ground truth and predicted translations ($\mathbf{t}_{gt}, \mathbf{t}$) as: 
\begin{equation}
\vspace{-0.25cm}
 \mathbf{\varphi} = \cos^{-1}\left(
 0.5(\text{tr}\left(\mathbf{R}_{gt}^T\mathbf{R}\right) - 1)\right),\,\, 
 \mathbf{\Delta t} = \lVert\mathbf{t}_{gt} - \mathbf{t}\rVert.
\end{equation}

\vspace{0.1cm}
\section{Experiments and Results}\label{sec:ExperimentalResults}
\vspace{-0.15cm}
\subsection{Indoor Scenes: Synthetic ModelNet40}\label{subsec:On_ModelNet40}
\vspace{-0.15cm}
\begin{figure}[!ht]
\begin{center}
\includegraphics[width=3.0in]{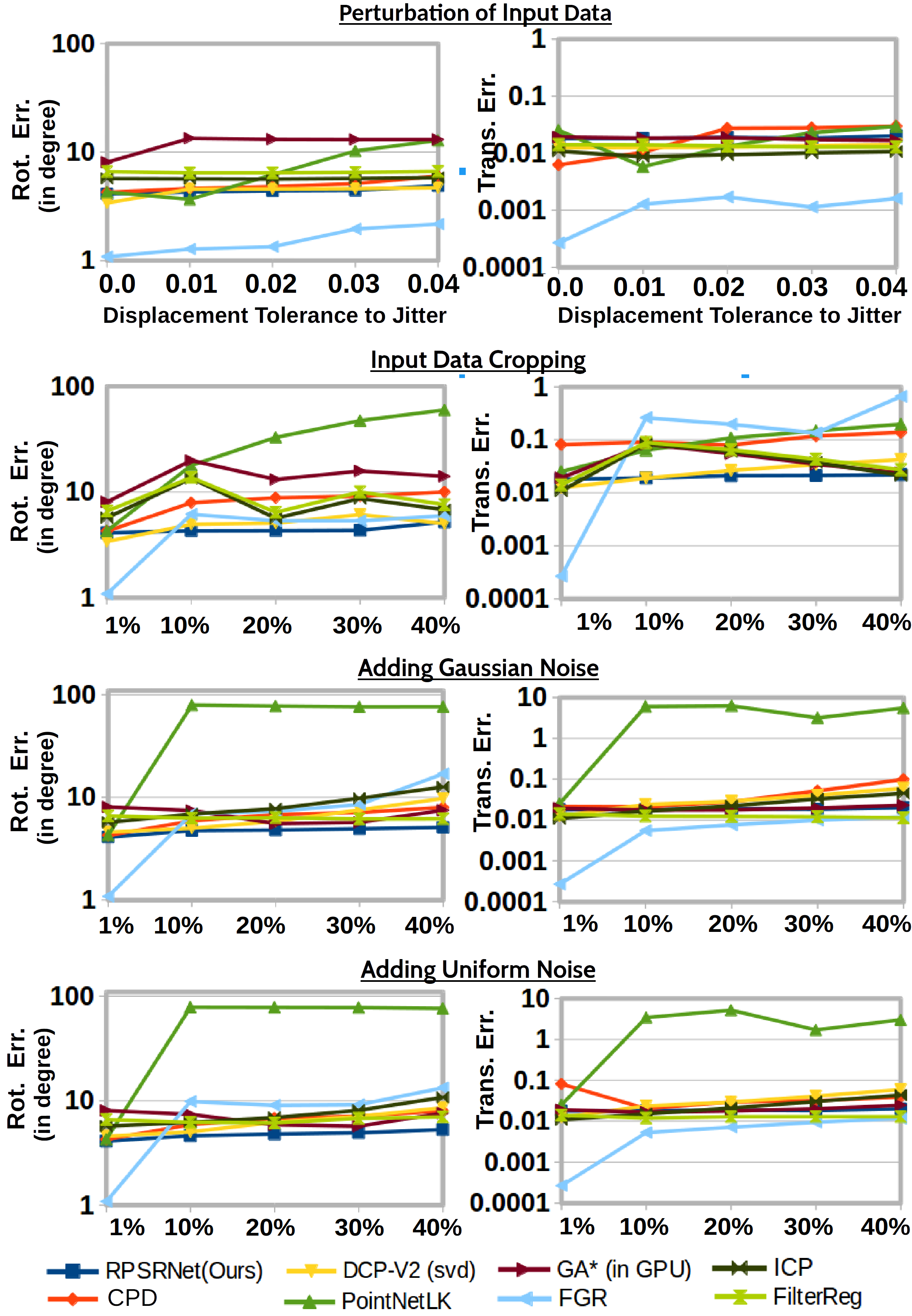} 
\end{center}
\vspace{-0.25cm}
\caption{Transformation error evaluated by DCP-v2~\cite{wang2019deep}, PointNetLK~\cite{AokiGSL19}, GA~\cite{Golyanik2016}, FGR~\cite{FGReccv16}, FilterReg~\cite{Gao2019} , ICP~\cite{BeslMcKay1992} and our RPSRNet on corrupted ModelNet40~\cite{ModelNet40}. Five increasing level of disturbances for each of the four types of disturbances. RPSRNet is stable performer and clear winner for most of the test instances.}
\label{fig:ModelNetEvaluation}
\end{figure}
\begin{figure*}[!ht]
\begin{center}
\includegraphics[width=6.9in]{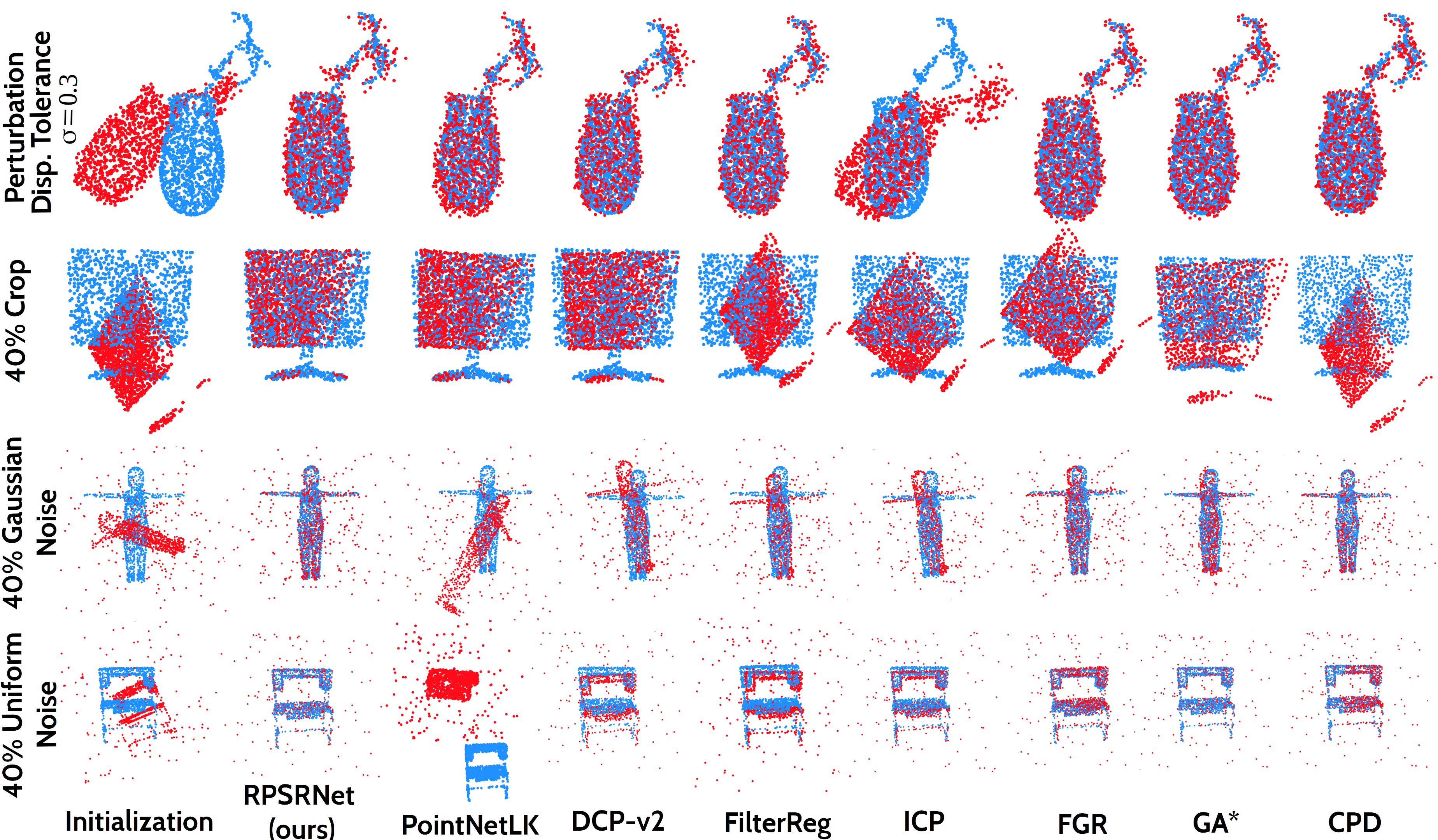}
\end{center}
\vspace{-0.25cm}
\caption{Qualitative registration outcomes on few samples fetched during evaluation
of competing methods against noisy data.}
\label{fig:ModelNet_SeenUnseen}
\end{figure*}

Since the DCP and PointNetLK provide the pre-trained model only on the clean version 
of the ModelNet40~\cite{ModelNet40} dataset, we retrain the networks on our augmented 
versions \textbf{(M1)} and \textbf{(M2)}.  
RPSRNet, DCP-v2~\cite{wang2019deep}, PointNetLK~\cite{AokiGSL19}, ICP~\cite{BeslMcKay1992},
CPD~\cite{MyronenkoSong2010}, FilterReg~\cite{Gao2019}, FGR~\cite{FGReccv16}, and 
GA~\cite{Golyanik2016} are evaluated on all twenty validation sets (for each different 
type and level of data disturbances).  
Despite training with the additional $950$ samples, both DCP and PointNetLK show a 
common generalizability issue. The error plots in Fig.~\ref{fig:ModelNetEvaluation} 
show increasing nature prediction inaccuracies for DCP and PointNetLK, 
with the increasing level of data disturbances. 
For instance, the transformation error of PointNetLK jumps several times 
higher when the noise level increases from $1\%$ to $5\%$ -- for Gaussian noise, 
the rotational error $\varphi_{\text{rmse}}$ increases from $2.267^\circ$ to $82.69^\circ$ and 
the translational error $\mathbf{\Delta t}_{rmse}$ increases from $0.1844$ to $5.945$. 
The same increment also occurs for the Uniform noise.
When the input data is clean, FGR~\cite{FGReccv16} performs 
the best with the lowest transformation error ($\varphi_{\text{rmse}}$ =  $1.082^\circ$ and 
$\mathbf{\Delta t}_{rmse}$ = $0.000267$). 
FGR is also consistently superior to all other competing methods at every level of
data perturbation, but its translation error is significantly higher in case of partial 
data registration. 
With the increasing level of noise, FGR's performance deteriorates further from being 
the best (at $1\%$ noise level) to the second-worst (at $40\%$ noise level).  
DCP approach is more robust than PointNetLK but the noise intolerance issue is still pertinent for both methods. 
RPSRNet is far more robust and stable than the competing methods (Fig.~\ref{fig:ModelNet_SeenUnseen} shows qualitative results on few evaluation samples).
\begin{table*}
\setlength{\tabcolsep}{2pt}
\footnotesize
\begin{center}
\begin{tabular}{@{}c c|c|c|c|c|c|c|c|c@{}}
\specialrule{.125em}{.1em}{.1em} 
\multicolumn{10}{c}{On KITTI~\cite{Geiger2013IJRR} Dataset}\\
\specialrule{.1em}{.05em}{.05em} 
\textbf{Seq.}
& \shortstack{\textbf{CPD}~\cite{MyronenkoSong2010}\\ $\mathbf{\varphi}_{\text{rmse}}$, $\mathbf{\Delta t}_{\text{rmse}}$}
& \shortstack{\textbf{GA*}~\cite{Golyanik2016}\\ $\mathbf{\varphi}_{\text{rmse}}$, $\mathbf{\Delta t}_{\text{rmse}}$}
& \shortstack{\textbf{FGR}~\cite{FGReccv16}\\ $\mathbf{\varphi}_{\text{rmse}}$, $\mathbf{\Delta t}_{\text{rmse}}$}
& \shortstack{\textbf{ICP}~\cite{BeslMcKay1992}\\ $\mathbf{\varphi}_{\text{rmse}}$, $\mathbf{\Delta t}_{\text{rmse}}$}
& \shortstack{\textbf{FilterReg}~\cite{Gao2019}\\ $\mathbf{\varphi}_{\text{rmse}}$, $\mathbf{\Delta t}_{\text{rmse}}$}
& \shortstack{\textbf{DCP-v2}~\cite{wang2019deep}\\ $\mathbf{\varphi}_{\text{rmse}}$, $\mathbf{\Delta t}_{\text{rmse}}$}
& \shortstack{\textbf{PointNetLK}~\cite{AokiGSL19}\\ $\mathbf{\varphi}_{\text{rmse}}$, $\mathbf{\Delta t}_{\text{rmse}}$}
& \shortstack{\textbf{RPSRNet$^1$ (ours)}\\ $\mathbf{\varphi}_{\text{rmse}}$, $\mathbf{\Delta t}_{\text{rmse}}$}
& \shortstack{\textbf{RPSRNet$^3$ (ours)}\\ $\mathbf{\varphi}_{\text{rmse}}$, $\mathbf{\Delta t}_{\text{rmse}}$}
\\
%
%
\specialrule{.125em}{0em}{0em}
\multirow{2}{*}{\textbf{mean}}    
&3.55, 1.08 
&3.30, 1.0 
&3.29, 0.85 
&3.15, 1.08 
&3.08, 0.77 
&2.92, 0.89
&4.02, 1.12 
&3.13, 0.88 
&\textbf{2.22, 0.58}
\\ \cmidrule(lr){2-10}
&\textcolor{mygray}{\textsl{3.03, 1.07}} 
&\textcolor{mygray}{\textsl{2.94, 1.02}} 
&\textcolor{mygray}{\textsl{3.25, 1.11}} 
&\textcolor{mygray}{\textsl{3.06, 1.20}} 
&\textcolor{mygray}{\textsl{3.26, 1.20}} 
&\textcolor{mygray}{\textsl{2.96, \underline{\textbf{0.76}}}} 
&\textcolor{mygray}{\textsl{5.17, 1.20}} 
&\textcolor{mygray}{\textsl{3.03, 1.01}} 
&\textcolor{mygray}{\textsl{\underline{\textbf{2.18}}, 0.84}}\\
\bottomrule
\end{tabular}
\end{center}
\caption{Evaluation on KITTI~\cite{Geiger2013IJRR} LiDAR sequences 00 to 07. Each cell in the table denotes the RMSE on angular and translational deviations from ground-truth when averaged over all 8 sequences. The nested-rows (upper and lower) denote the \textbf{(K1-w/o)} and \textcolor{mygray}{\textbf{(K2-w)}} setups. The lowest transformation errors achieved by any method is highlighted in \textbf{bold} or underlined \textcolor{mygray}{\underline{\textbf{\textsl{bold}}}} font.}
\label{table:Ablation_RPSRNet_KITTI}
\end{table*}
\vspace{-0.15cm}
\begin{figure}[!ht]
\begin{center}
\includegraphics[width=3in]{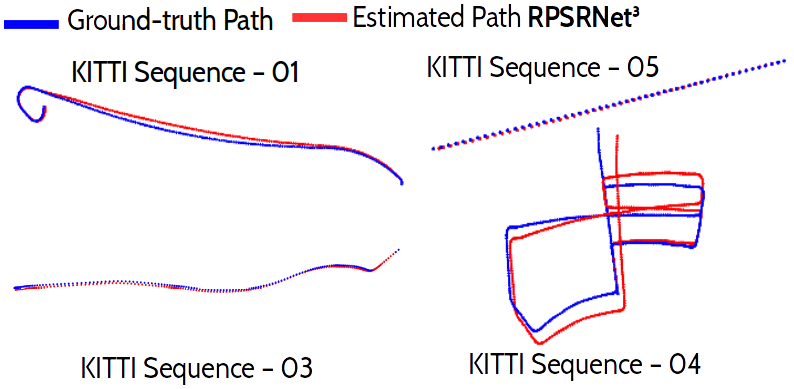} 
\end{center}
\vspace{-0.5cm}
\caption{Vehicle trajectories for the KITTI LiDAR sequences are obtained by registering consecutive pairs of frames.}
\label{fig:OdometryPath}
\end{figure}

\vspace{-0.25cm}
\subsection{Outdoor Scene: KITTI LiDAR Dataset}\label{subsec:On_KITTI}
\begin{figure}[!ht]
\begin{center}
\includegraphics[width=0.99\linewidth]{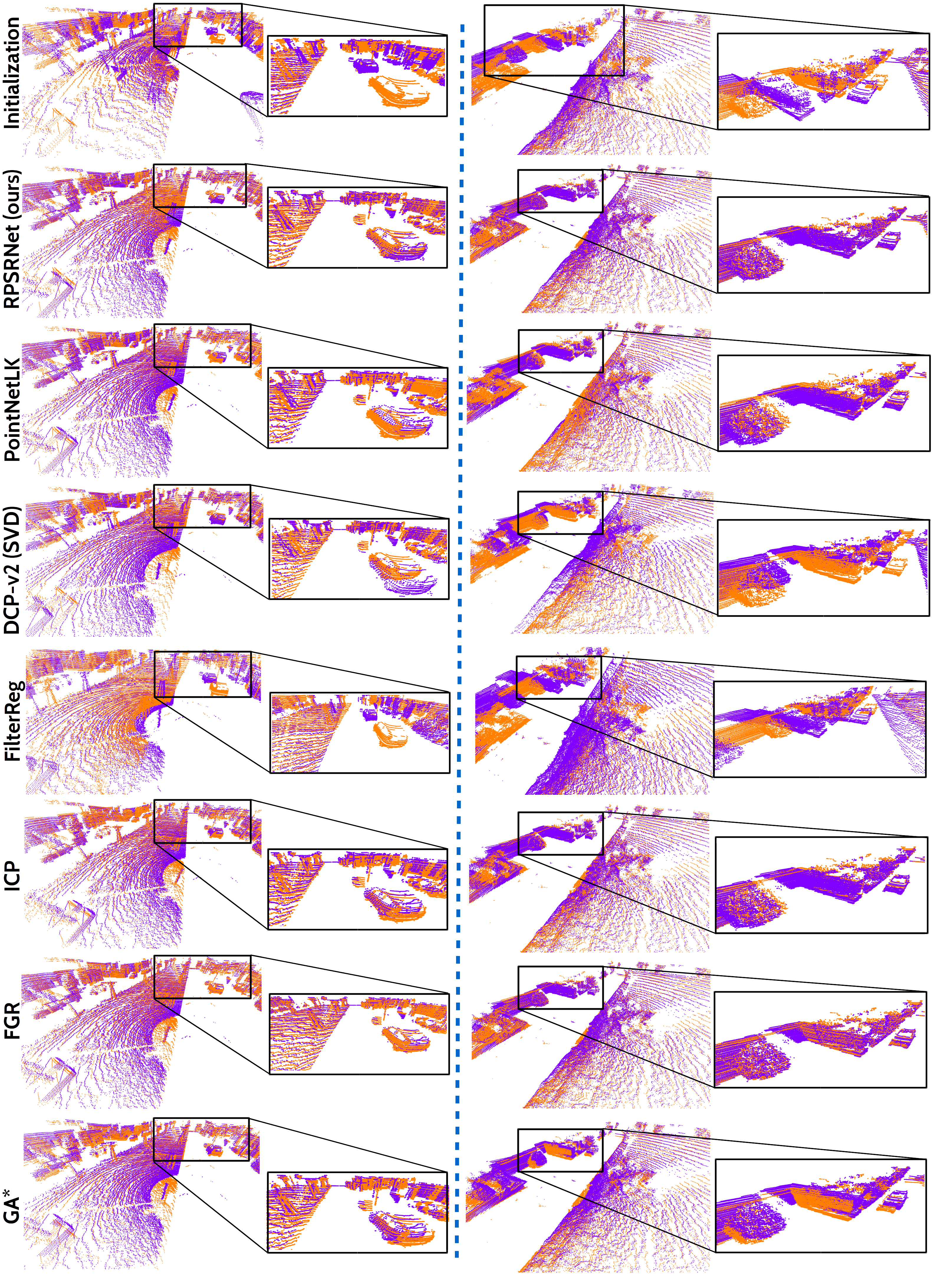} 
\end{center}
\vspace{-0.5cm}
\caption{Results of our RPSRNet$^3$ on some samples 
from the validation set of KITTI~\cite{Geiger2013IJRR, behley2019iccv} dataset (\textit{left column}: frame 000477 (as $\mathbf{Y}$) and 000482 (as $\mathbf{X}$) of seq-03 and \textit{right column}: frame 000019 (as $\mathbf{Y}$) and 000024 (as $\mathbf{X}$) from the seq-06). Zoomed parts of each image highlights how other competing methods perform in aligning static target objects -- \eg, cars and bushes.}
\label{fig:KITTI_Comparison}
\end{figure}

\vspace{-0.1cm}
Experiments show most unsupervised alignment methods ICP~\cite{BeslMcKay1992}, 
CPD~\cite{MyronenkoSong2010}, FilterReg~\cite{Gao2019}, and GA~\cite{Golyanik2016} all fail to recover mainly the correct translation difference between \textit{source} 
and \textit{target}, and gets trapped into bad local minima. 
The evaluation Table~\ref{table:Ablation_RPSRNet_KITTI}, reports the final RMSE values on orientations and translations averaged\footnote{The supplementary document provides a detailed evaluation on the individual sequences} over \textit{eight} sequences (00 - 07) for all baseline methods. 
The upper and lower sub-rows indicate \textbf{(K1-w/o)} and \textcolor{mygray}{\textbf{(K2-w)}} setups respectively.  
Like in ModelNet40 experiment, FGR~\cite{FGReccv16} performs better among the unsupervised approaches.
Our RPSRNet (the last two columns) outperforms all competing methods. 
For instance, in \textcolor{mygray}{\textbf{K2-w}} setup, unsupervided methods ICP, FilterReg, CPD and GA reports $\Delta t_{\text{rmse}}$ as 1.08, 0.77, 1.08, and 1.0 which are 1.87, 1.33, 1.87, and 1.72 times higher than RPSRNet's error value 0.58.
In the \textbf{K1-w/o} setup, all methods record higher transformation errors, especially on translation part, than RPSRNet. 
A small rotational inaccuracy in range of $0.5^\circ$ to $3^\circ$ after registration is acceptable because further refinement using ICP or other fast alignment \cite{2020arXiv200914005A} methods reduce such difference. 
But a prediction error beyond $70\,\text{cm}$ for the translation part denotes large dispute. 
In \textbf{K1-w/o} setup: RPSRNet$^3$ has 1.31 and 1.33 times lower rotational and translational errors than the respective second best methods DCP-v2 and FilterReg. 
On the other hand, in \textcolor{mygray}{\textbf{K2-w}} setup: the same error for rotation is 1.35 lower than second best candidate GA~\cite{Golyanik2016}, but the translation error is 1.1 times higher as the second best behind DCP-v2.
The Fig.~\ref{fig:KITTI_Comparison} shows some qualitative results from our RPSRNet$^3$ compared to other competing methods on some challenging frames from sequence 03 and 06. 
%
%
%
%

\vspace{0.05cm}
\noindent\textbf{Frame-to-Frame LiDAR Registration.} 
Using the trained model of RPSRNet, we predict the relative transformations 
on all consecutive pairs of frames in a given sequence. 
Lets assume that the relative transformation between the sensor poses from frame $f$ to $f+1$ is $\mathbf{T}_{4\times4}^f=\left[\mathbf{R}_f\,\mathbf{t}_{f}\right]$ \hbox{w.r.t} the initial sensor pose $\left[\mathbf{R_{\text{init}},t_{\text{init}}}\right]$.
The trajectory of a point $\vec{p} = (0, 0, 0, 1)^T$ shown in the Fig.~\ref{fig:OdometryPath} is the locus of its spatial positions starting from frame $1$ as $\vec{p}^1$ till the frame $f$ as
\vspace{-0.15cm}
\begin{equation}
\small
\vec{p}^f = \left[\mathbf{R}_{\text{init}} \mathbf{t}_{\text{init}}\right]\cdot 
\left(\left[\mathbf{R}_f\, \mathbf{t}_{f}\right] 
      \cdot 
      \left[\mathbf{R}_{f-1}\, \mathbf{t}_{f-1}\right]
      \cdots
      \left[\mathbf{R}_1\, \mathbf{t}_{1}\right]
\right)^{-1} \cdot \vec{p}.
\vspace{-0.05cm}
\end{equation}for four different sequences. Our measurements are close to the ground-truth.

\vspace{0.1cm}
\subsection{Runtime and Efficiency}\label{subsec:RuntimeAnalysis}
\begin{figure}[!ht]
\begin{center}
\includegraphics[width=3.25in]{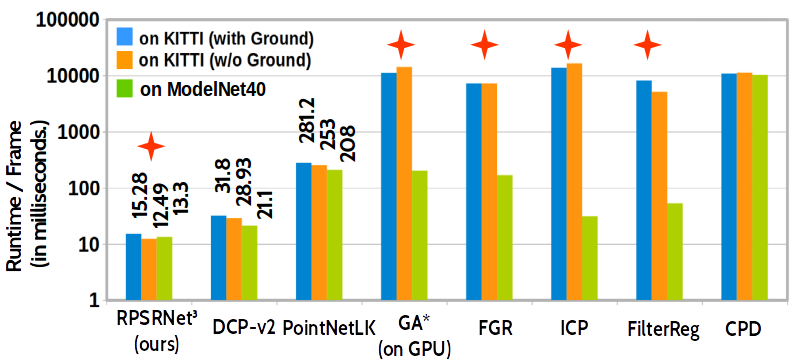} 
\end{center}
\vspace{-0.25cm}
\caption{
Runtime of the competing methods on KITTI and ModelNet40 with values for 
the deep-learning based methods. The methods with star marks above take the 
whole point cloud, whereas others use subsampled version of ${\sim}$2K.}
\label{fig:RuntimeRPSRNet}
\end{figure}

\vspace{-0.1cm}
RPSRNet is computationally efficient with the fastest inference time among the competing methods on all tested datasets (see Fig.~\ref{fig:RuntimeRPSRNet}).
RPSRNet, FGR~\cite{FGReccv16}, ICP~\cite{BeslMcKay1992}, FilterReg~\cite{Gao2019} and GA~\cite{Golyanik2016} takes the input with its actual point size for KITTI whereas other methods -- DCP-v2~\cite{wang2019deep}, PointNetLK~\cite{AokiGSL19} and CPD~\cite{MyronenkoSong2010} takes downsampled version (2048 points / frame) of the point clouds otherwise special GPU or CPU memory cards are required.
The input point size (2048 points / sample) is constant for all methods in case of ModelNet40 dataset. 
RPSRNet, DCP-v2, PointNetLK, GA run on an NVIDIA Titan 1080 GPU, whereas CPD, FGR, ICP and FilterReg are run on a 3.0GHz Intel Xeon CPU. 
Training time per epoch for RPSRNet, DCP-v2, PointNetLK are 5, 18, and 12 minutes, respectively. 

\vspace{-0.20cm}

\subsection{Ablation Study}\label{sec:Ablation_RPSRNet}
\vspace{-0.15cm}
\begin{figure}[!ht]
\begin{center}
\includegraphics[width=3.3in]{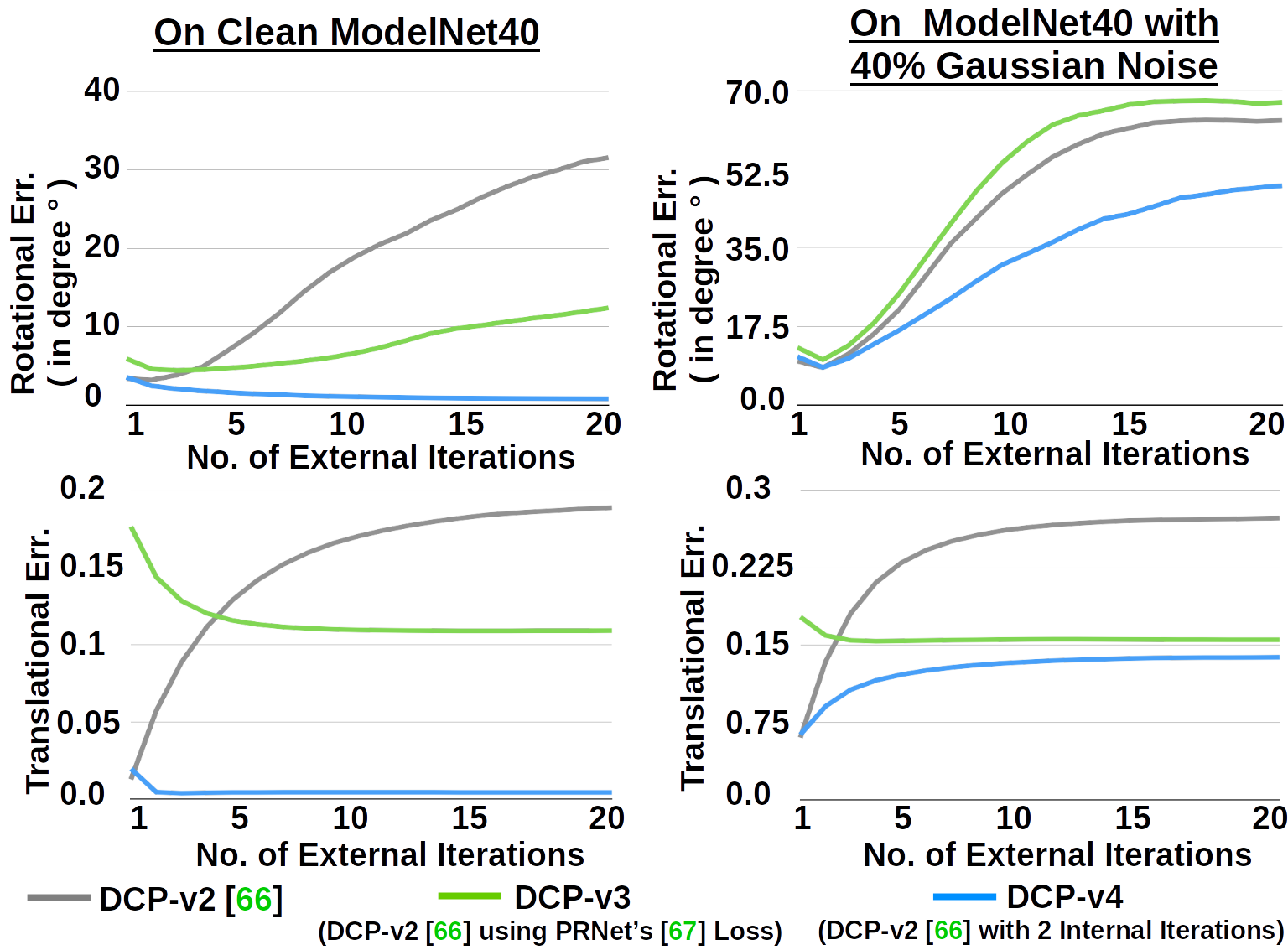} 
\end{center}
\vspace{-0.25cm}
\caption{On clean ModelNet40~\cite{ModelNet40} (\textit{left}) : If we use iterative refinement, 
the transformation errors of DCP-v2~\cite{wang2019deep} and DCP-v3 (modified version of 
DCP-v2 including additional loss functions from PRNet~\cite{wang2019prnet}) start diverging 
to increase. The same observation also found on ModelNet40 with $40\%$ Gaussian noise for all three methods 
(\textit{right}).}
\label{fig:DCPv2Ablation}
\end{figure}
\noindent\textbf{Why not DCP-v2 with iterative refinement? } 
In this ablation, we implement two hybrid versions of DCP-v2 -- (i) DCP-v3: it uses DCP-v2 architecture with additional loss functions (cyclic consistency and embedding disparity) from its extension PRNet~\cite{wang2019prnet} and (ii) DCP-v4: this is DCP-v2 with two internal alignment network blocks. Fig.~\ref{fig:DCPv2Ablation} shows that DCP-v2 and DCP-v3 diverge when tested on clean ModelNet40 with increasing number of external iterations for prediction refinement. 
On noisy version of ModelNet40, all three versions of DCP (v2, v3, v4) diverge with increasing errors. This test states that an iterative approach similar to us with additional loss functions does not help DCP-v2~\cite{wang2019deep}. It also reasserts how our novel representation, $2^D$-tree convolution, and weighted feature embedding improve the overall accuracy.

\vspace{-0.2cm}

\section{Conclusions and Discussion}
\vspace{-0.15cm}
Our work presents a novel input representation and an end-to-end learning framework for rigid point set registration using $2^3$-tree convolution. 
Extensive experiments on different types of datasets show RPSRNet outperforms competing methods in terms of inference speed and accuracy.
The proposed network has limitations to tackle partial-to-partial data registration which is a higher level challenge.  

We plan to integrate loss functions which deal with partially overlapping data, \eg, Chamfer distance, and also extend our network architecture for estimating non-rigid displacement fields, \eg scene-flow. 
Hence, in future, we will implement the deconvolution blocks against each layer of the encoder part in the current network. 
At present, we build the BH-trees of the input point clouds on CPU (in 4-5 ms./sample) before training starts and pass its attributes to the network during training 
time as arrays. It disrupts a single application flow. 
We will integrate a GPU implementation of BH-Tree~\cite{BURTSCHER201175} for direct 
input data processing. 

{\small
\bibliographystyle{ieee_fullname}
\bibliography{egbib}

\begin{thebibliography}{10}\itemsep=-1pt

\bibitem{Agarwal20173DPC}
Swapna Agarwal and Brojeshwar Bhowmick.
\newblock 3d point cloud registration with shape constraint.
\newblock {\em International Conference on Image Processing (ICIP)}, pages
  2199--2203, 2017.

\bibitem{aldoma2011cad}
Aitor Aldoma, Markus Vincze, Nico Blodow, David Gossow, Suat Gedikli,
  Radu~Bogdan Rusu, and Gary Bradski.
\newblock Cad-model recognition and 6dof pose estimation using 3d cues.
\newblock In {\em International Conference on Computer Vision (ICCV)
  Workshops}, 2011.

\bibitem{Ali2018NRGAGA}
Sk~Aziz Ali, Vladislav Golyanik, and Didier Stricker.
\newblock {NRGA: Gravitational Approach for Non-rigid Point Set Registration}.
\newblock In {\em International Conference on 3D Vision (3DV)}, 2018.

\bibitem{2020arXiv200914005A}
Sk~Aziz {Ali}, Kerem {Kahraman}, Christian {Theobalt}, Didier {Stricker}, and
  Vladislav {Golyanik}.
\newblock {Fast Gravitational Approach for Rigid Point Set Registration with
  Ordinary Differential Equations}.
\newblock {\em arXiv e-prints}, 2020.

\bibitem{AokiGSL19}
Yasuhiro Aoki, Hunter Goforth, Rangaprasad~Arun Srivatsan, and Simon Lucey.
\newblock Pointnetlk: Robust and efficient point cloud registration using
  pointnet.
\newblock In {\em CVPR}, 2019.

\bibitem{Transformer2017}
Noam~Shazeer Ashish~Vaswani, Niki Parmar, Jakob Uszkoreit, Llion Jones, Aidan~N
  Gomez, Lukasz Kaiser, , and Illia Polo-sukhin.
\newblock Attention is all you need.
\newblock In {\em Advances in Neural Information Processing (NIPS)}, 2017.

\bibitem{bader2012space}
Michael Bader.
\newblock {\em Space-filling curves: an introduction with applications in
  scientific computing}, volume~9.
\newblock Springer Science \& Business Media, 2012.

\bibitem{1986Natur324446B}
J. Barnes and P. Hut.
\newblock A hierarchical o(n log n) force-calculation algorithm.
\newblock {\em Nature}, 324:446--449, 1986.

\bibitem{behley2019iccv}
J. Behley, M. Garbade, A. Milioto, J. Quenzel, S. Behnke, C. Stachniss, and J.
  Gall.
\newblock {SemanticKITTI: A Dataset for Semantic Scene Understanding of LiDAR
  Sequences}.
\newblock In {\em International Conf.~on Computer Vision (ICCV)}, 2019.

\bibitem{BeslMcKay1992}
P.~J. {Besl} and N.~D. {McKay}.
\newblock A method for registration of 3-d shapes.
\newblock {\em Transactions on Pattern Analysis and Machine Intelligence
  (TPAMI)}, 14(2):239--256, 1992.

\bibitem{GMMReg1544863}
{Bing Jian} and B.~C. {Vemuri}.
\newblock A robust algorithm for point set registration using mixture of
  gaussians.
\newblock In {\em International Conference on Computer Vision (ICCV)}, 2005.

\bibitem{BURTSCHER201175}
Martin Burtscher and Keshav Pingali.
\newblock Chapter 6 - an efficient cuda implementation of the tree-based barnes
  hut n-body algorithm.
\newblock In {\em GPU Computing Gems Emerald Edition}, Applications of GPU
  Computing Series, pages 75 -- 92. Boston, 2011.

\bibitem{choy2020deep}
Christopher Choy, Wei Dong, and Vladlen Koltun.
\newblock Deep global registration.
\newblock In {\em Computer Vision and Pattern Recognition (CVPR)}, 2020.

\bibitem{choy2020high}
Christopher Choy, Junha Lee, Rene Ranftl, Jaesik Park, and Vladlen Koltun.
\newblock High-dimensional convolutional networks for geometric pattern
  recognition.
\newblock In {\em Computer Vision and Pattern Recognition (CVPR)}, 2020.

\bibitem{choy2019fully}
Christopher Choy, Jaesik Park, and Vladlen Koltun.
\newblock Fully convolutional geometric features.
\newblock In {\em International Conference on Computer Vision}, pages
  8958--8966, 2019.

\bibitem{DengCVPR2019}
H. {Deng}, T. {Birdal}, and S. {Ilic}.
\newblock 3d local features for direct pairwise registration.
\newblock In {\em Computer Vision and Pattern Recognition (CVPR)}, 2019.

\bibitem{devlin2018bert}
Jacob Devlin, Ming-Wei Chang, Kenton Lee, and Kristina Toutanova.
\newblock Bert: Pre-training of deep bidirectional transformers for language
  understanding.
\newblock {\em arXiv preprint arXiv:1810.04805}, 2018.

\bibitem{BSC2017}
Zhen Dong, Bisheng Yang, Yuan Liu, Fuxun Liang, Bijun Li, and Yufu Zang.
\newblock A novel binary shape context for 3d local surface description.
\newblock 130:431 -- 452, 2017.

\bibitem{Eckart2018HGMRHG}
Benjamin Eckart, Kihwan Kim, and Jan Kautz.
\newblock Hgmr: Hierarchical gaussian mixtures for adaptive 3d registration.
\newblock In {\em European Conference on Computer Vision (ECCV)}, 2018.

\bibitem{Elbaz20173DPC}
Gil Elbaz, Tamar Avraham, and Anath Fischer.
\newblock 3d point cloud registration for localization using a deep neural
  network auto-encoder.
\newblock {\em Computer Vision and Pattern Recognition (CVPR)}, pages
  2472--2481, 2017.

\bibitem{Gao2019}
Wei Gao and Russ Tedrake.
\newblock Filterreg: Robust and efficient probabilistic point-set registration
  using gaussian filter and twist parameterization.
\newblock In {\em Computer Vision and Pattern Recognition (CVPR)}, 2019.

\bibitem{Geiger2013IJRR}
Andreas Geiger, Philip Lenz, Christoph Stiller, and Raquel Urtasun.
\newblock Vision meets robotics: The kitti dataset.
\newblock {\em International Journal of Robotics Research (IJRR)}, 2013.

\bibitem{GeomStableSamplingICP2003}
N. {Gelfand}, L. {Ikemoto}, S. {Rusinkiewicz}, and M. {Levoy}.
\newblock Geometrically stable sampling for the icp algorithm.
\newblock In {\em International Conference on 3-D Digital Imaging and Modeling,
  2003. 3DIM 2003. Proceedings.}, pages 260--267, 2003.

\bibitem{gojcic2019perfect}
Zan Gojcic, Caifa Zhou, Jan~D Wegner, and Andreas Wieser.
\newblock The perfect match: 3d point cloud matching with smoothed densities.
\newblock In {\em Computer Vision and Pattern Recognition (CVPR)}, pages
  5545--5554, 2019.

\bibitem{Golyanik2016}
Vladislav Golyanik, Sk~Aziz Ali, and Didier Stricker.
\newblock Gravitational approach for point set registration.
\newblock {\em Computer Vision and Pattern Recognition (CVPR)}, 2016.

\bibitem{BHRGA2019}
Vladislav Golyanik, Christian Theobalt, and Didier Stricker.
\newblock Accelerated gravitational point set alignment with altered physical
  laws.
\newblock In {\em International Conference on Computer Vision (ICCV)}, 2019.

\bibitem{gower1975generalized}
John~C Gower.
\newblock Generalized procrustes analysis.
\newblock {\em Psychometrika}, 40(1):33--51, 1975.

\bibitem{EM-ICP_Granger}
S{\'e}bastien Granger and Xavier Pennec.
\newblock Multi-scale em-icp: A fast and robust approach for surface
  registration.
\newblock In {\em European Conference on Computer Vision (ECCV)}. Springer
  Berlin Heidelberg, 2002.

\bibitem{Greenspan2003}
MA Greenspan and M. Yurick.
\newblock Approximate k-d tree search for efficient icp.
\newblock In {\em International Conference on Recent Advances in 3D Digital
  Imaging and Modeling (3DIM)}, 2003.

\bibitem{gross2019alignnet}
Johannes Gro{\ss}, Aljo{\v{s}}a O{\v{s}}ep, and Bastian Leibe.
\newblock Alignnet-3d: Fast point cloud registration of partially observed
  objects.
\newblock In {\em International Conference on 3D Vision (3DV)}, pages 623--632.
  IEEE, 2019.

\bibitem{DeepResNet2016}
K. {He}, X. {Zhang}, S. {Ren}, and J. {Sun}.
\newblock Deep residual learning for image recognition.
\newblock In {\em Computer Vision and Pattern Recognition (CVPR)}, 2016.

\bibitem{Huang_2020_CVPR}
Lila Huang, Shenlong Wang, Kelvin Wong, Jerry Liu, and Raquel Urtasun.
\newblock Octsqueeze: Octree-structured entropy model for lidar compression.
\newblock In {\em Computer Vision and Pattern Recognition (CVPR)}, 2020.

\bibitem{BatchNorm2015}
Sergey Ioffe and Christian Szegedy.
\newblock Batch normalization: Accelerating deep network training by reducing
  internal covariate shift.
\newblock In {\em International Conference on Machine Learning (ICML)}, 2015.

\bibitem{Kabsch1976}
Wolfgang Kabsch.
\newblock A solution for the best rotation to relate two sets of vectors.
\newblock {\em Acta Crystallographica Section A: Crystal Physics, Diffraction,
  Theoretical and General Crystallography}, 32(5):922--923, 1976.

\bibitem{Khoury2017LearningCG}
Marc Khoury, Qian-Yi Zhou, and Vladlen Koltun.
\newblock Learning compact geometric features.
\newblock In {\em International Conference on Computer Vision (ICCV)}, pages
  153--161, 2017.

\bibitem{Korn2014ColorSG}
Michael Korn, Martin Holzkothen, and Josef Pauli.
\newblock Color supported generalized-icp.
\newblock {\em Computer Vision Theory and Applications (VISAPP)}, 3:592--599,
  2014.

\bibitem{lei2019octree}
Huan Lei, Naveed Akhtar, and Ajmal Mian.
\newblock Octree guided cnn with spherical kernels for 3d point clouds.
\newblock {\em Computer Vision and Pattern Recognition (CVPR)}, 2019.

\bibitem{li2019net}
Qing Li, Shaoyang Chen, Cheng Wang, Xin Li, Chenglu Wen, Ming Cheng, and
  Jonathan Li.
\newblock Lo-net: Deep real-time lidar odometry.
\newblock In {\em Computer Vision and Pattern Recognition (CVPR)}, 2019.

\bibitem{s19194188}
Weiping Liu, Jia Sun, Wanyi Li, Ting Hu, and Peng Wang.
\newblock Deep learning on point clouds and its application: A survey.
\newblock {\em Sensors}, 19(19), 2019.

\bibitem{liu2019densepoint}
Yongcheng Liu, Bin Fan, Gaofeng Meng, Jiwen Lu, Shiming Xiang, and Chunhong
  Pan.
\newblock Densepoint: Learning densely contextual representation for efficient
  point cloud processing.
\newblock In {\em International Conference on Computer Vision (ICCV)}, 2019.

\bibitem{liu2019rscnn}
Yongcheng Liu, Bin Fan, Shiming Xiang, and Chunhong Pan.
\newblock Relation-shape convolutional neural network for point cloud analysis.
\newblock In {\em Computer Vision and Pattern Recognition (CVPR)}, 2019.

\bibitem{lu2019deepvcp}
Weixin Lu, Guowei Wan, Yao Zhou, Xiangyu Fu, Pengfei Yuan, and Shiyu Song.
\newblock Deepvcp: An end-to-end deep neural network for point cloud
  registration.
\newblock In {\em Computer Vision and Pattern Recognition (CVPR)}, pages
  12--21, 2019.

\bibitem{maturana2015voxnet}
Daniel Maturana and Sebastian Scherer.
\newblock Voxnet: A 3d convolutional neural network for real-time object
  recognition.
\newblock In {\em 2015 IEEE/RSJ International Conference on Intelligent Robots
  and Systems (IROS)}, pages 922--928. IEEE.

\bibitem{MyronenkoSong2010}
A. {Myronenko} and X. {Song}.
\newblock Point set registration: Coherent point drift.
\newblock {\em Transactions on Pattern Analysis and Machine Intelligence
  (TPAMI)}, 32(12):2262--2275, 2010.

\bibitem{ReLuVinod2010}
Vinod Nair and Geoffrey~E. Hinton.
\newblock Rectified linear units improve restricted boltzmann machines.
\newblock In {\em International Conference on Machine Learning (ICML)}, 2010.

\bibitem{KinextFusion_Newcombe}
R.~A. {Newcombe}, S. {Izadi}, O. {Hilliges}, D. {Molyneaux}, D. {Kim}, A.~J.
  {Davison}, P. {Kohi}, J. {Shotton}, S. {Hodges}, and A. {Fitzgibbon}.
\newblock Kinectfusion: Real-time dense surface mapping and tracking.
\newblock In {\em International Symposium on Mixed and Augmented Reality
  (ISMAR)}, 2011.

\bibitem{Nezhadarya_2020_CVPR}
Ehsan Nezhadarya, Ehsan Taghavi, Ryan Razani, Bingbing Liu, and Jun Luo.
\newblock Adaptive hierarchical down-sampling for point cloud classification.
\newblock In {\em Conference on Computer Vision and Pattern Recognition
  (CVPR)}, June.

\bibitem{3DRegNetCVPR19}
G.~Dias Pais, Pedro Miraldo, Srikumar Ramalingam, Jacinto~C. Nascimento,
  Venu~Madhav Govindu, and Rama Chellappa.
\newblock 3dregnet: A deep neural network for 3d point registration.
\newblock 2019.

\bibitem{AutoDiffPytorch}
Adam Paszke, Sam Gross, Soumith Chintala, Gregory Chanan, Edward Yang, Zachary
  DeVito, Zeming Lin, Alban Desmaison, Luca Antiga, , and Adam Lerer.
\newblock Automatic differentiation in pytorch. 2017.

\bibitem{Pomerleau2013}
Francis Pomerleau, Fran{\c{c}}oisand~Colas and St{\'e}phane Siegwart,
  Rolandand~Magnenat.
\newblock Comparing icp variants on real-world data sets.
\newblock {\em Autonomous Robots}, 34(3):133--148, 2013.

\bibitem{qi2017pointnet}
Charles~R Qi, Hao Su, Kaichun Mo, and Leonidas~J Guibas.
\newblock Pointnet: Deep learning on point sets for 3d classification and
  segmentation.
\newblock In {\em Computer Vision and Pattern Recognition (CVPR)}, pages
  652--660, 2017.

\bibitem{qi2017pointnet++}
Charles~Ruizhongtai Qi, Li Yi, Hao Su, and Leonidas~J Guibas.
\newblock Pointnet++: Deep hierarchical feature learning on point sets in a
  metric space.
\newblock In {\em Advances in neural information processing systems}, pages
  5099--5108, 2017.

\bibitem{raguram2008comparative}
Rahul Raguram, Jan-Michael Frahm, and Marc Pollefeys.
\newblock A comparative analysis of ransac techniques leading to adaptive
  real-time random sample consensus.
\newblock In {\em European Conference on Computer Vision (ECCV)}, 2008.

\bibitem{riegler2017octnet}
Gernot Riegler, Ali Osman~Ulusoy, and Andreas Geiger.
\newblock Octnet: Learning deep 3d representations at high resolutions.
\newblock In {\em Computer Vision and Pattern Recognition (CVPR)}, pages
  3577--3586, 2017.

\bibitem{Riegler2017CVPR}
Gernot Riegler, Osman Ulusoy, and Andreas Geiger.
\newblock Octnet: Learning deep 3d representations at high resolutions.
\newblock In {\em Computer Vision and Pattern Recognition (CVPR)}, 2017.

\bibitem{Rusinkiewicz2001}
Szymon Rusinkiewicz and Marc Levoy.
\newblock Efficient variants of the {ICP} algorithm.
\newblock In {\em International Conference on 3D Digital Imaging and Modeling
  (3DIM)}, 2001.

\bibitem{Rusu2009}
R.~B. {Rusu}, N. {Blodow}, and M. {Beetz}.
\newblock Fast point feature histograms (fpfh) for 3d registration.
\newblock In {\em International Conference on Robotics and Automation (ICRA)},
  2009.

\bibitem{Rusu2008}
R.~B. {Rusu}, N. {Blodow}, Z.~C. {Marton}, and M. {Beetz}.
\newblock Aligning point cloud views using persistent feature histograms.
\newblock In {\em International Conference on Intelligent Robots and Systems
  (IROS)}, 2008.

\bibitem{santoro2017simple}
Adam Santoro, David Raposo, David~G Barrett, Mateusz Malinowski, Razvan
  Pascanu, Peter Battaglia, and Timothy Lillicrap.
\newblock A simple neural network module for relational reasoning.
\newblock In {\em Advances in neural information processing systems}, pages
  4967--4976, 2017.

\bibitem{GeneralizedICPSegal}
Aleksandr Segal, Dirk H{\"{a}}hnel, and Sebastian Thrun.
\newblock Generalized-icp.
\newblock In {\em Robotics: Science and Systems V, University of Washington,
  Seattle, USA, June 28 - July 1, 2009}, 2009.

\bibitem{su18splatnet}
Hang Su, Varun Jampani, Deqing Sun, Subhransu Maji, Evangelos Kalogerakis,
  Ming-Hsuan Yang, and Jan Kautz.
\newblock {SPLATN}et: Sparse lattice networks for point cloud processing.
\newblock In {\em Computer Vision and Pattern Recognition (CVPR)}, 2018.

\bibitem{tatarchenko2017octree}
Maxim Tatarchenko, Alexey Dosovitskiy, and Thomas Brox.
\newblock Octree generating networks: Efficient convolutional architectures for
  high-resolution 3d outputs.
\newblock In {\em International Conference on Computer Vision}, 2017.

\bibitem{Umeyama1991}
S. {Umeyama}.
\newblock Least-squares estimation of transformation parameters between two
  point patterns.
\newblock {\em Transactions on Pattern Analysis and Machine Intelligence
  (TPAMI)}, 13(4):376--380, 1991.

\bibitem{OctreeNet2020}
F. {Wang}, Y. {Zhuang}, H. {Gu}, and H. {Hu}.
\newblock Octreenet: A novel sparse 3-d convolutional neural network for
  real-time 3-d outdoor scene analysis.
\newblock {\em IEEE Transactions on Automation Science and Engineering},
  17(2):735--747, 2020.

\bibitem{Wang-2017-ocnn}
Peng-Shuai Wang, Yang Liu, Yu-Xiao Guo, Chun-Yu Sun, and Xin Tong.
\newblock {O-CNN: Octree-based Convolutional Neural Networks for 3D Shape
  Analysis}.
\newblock {\em ACM Transactions on Graphics (SIGGRAPH)}, 36(4), 2017.

\bibitem{wang2019deep}
Yue Wang and Justin Solomon.
\newblock Deep closest point: Learning representations for point cloud
  registration.
\newblock In {\em International Conference on Computer Vision (ICCV)}, pages
  3523--3532, 2019.

\bibitem{wang2019prnet}
Yue Wang and Justin Solomon.
\newblock Prnet: Self-supervised learning for partial-to-partial registration.
\newblock In {\em Advances in Neural Information Processing Systems (NIPS)},
  pages 8812--8824, 2019.

\bibitem{wang2019dynamic}
Yue Wang, Yongbin Sun, Ziwei Liu, Sanjay~E Sarma, Michael~M Bronstein, and
  Justin~M Solomon.
\newblock Dynamic graph cnn for learning on point clouds.
\newblock {\em ACM Transactions on Graphics (TOG)}, 38(5):1--12, 2019.

\bibitem{wu2019pointconv}
Wenxuan Wu, Zhongang Qi, and Li Fuxin.
\newblock Pointconv: Deep convolutional networks on 3d point clouds.
\newblock In {\em Computer Vision and Pattern Recognition}, pages 9621--9630,
  2019.

\bibitem{Yang2013}
J. {Yang}, H. {Li}, and Y. {Jia}.
\newblock Go-icp: Solving 3d registration efficiently and globally optimally.
\newblock In {\em International Conference on Computer Vision (ICCV)}, 2013.

\bibitem{RPMNet2020}
Z.~J. {Yew} and G.~H. {Lee}.
\newblock Rpm-net: Robust point matching using learned features.
\newblock In {\em 2020 IEEE/CVF Conference on Computer Vision and Pattern
  Recognition (CVPR)}, 2020.

\bibitem{yifan2019patch}
Wang Yifan, Shihao Wu, Hui Huang, Daniel Cohen-Or, and Olga Sorkine-Hornung.
\newblock Patch-based progressive 3d point set upsampling.
\newblock In {\em Computer Vision and Pattern Recognition (CVPR)}, pages
  5958--5967, 2019.

\bibitem{yuan2020deepgmr}
Wentao Yuan, Benjamin Eckart, Kihwan Kim, Varun Jampani, Dieter Fox, and Jan
  Kautz.
\newblock Deepgmr: Learning latent gaussian mixture models for registration.
\newblock In {\em European Conference on Computer Vision (ECCV)}. Springer,
  2020.

\bibitem{Zhang-2016-110808}
Ji Zhang and Sanjiv Singh.
\newblock Low-drift and real-time lidar odometry and mapping.
\newblock {\em Autonomous Robots}, 41(2):401–416, 2016.

\bibitem{ModelNet40}
{Zhirong Wu}, S. {Song}, A. {Khosla}, {Fisher Yu}, {Linguang Zhang}, {Xiaoou
  Tang}, and J. {Xiao}.
\newblock 3d shapenets: A deep representation for volumetric shapes.
\newblock In {\em Computer Vision and Pattern Recognition (CVPR)}, 2015.

\bibitem{zhou2010data}
Kun Zhou, Minmin Gong, Xin Huang, and Baining Guo.
\newblock Data-parallel octrees for surface reconstruction.
\newblock {\em Transactions on visualization and computer graphics},
  17(5):669--681, 2010.

\bibitem{FGReccv16}
Qian-Yi Zhou, Jaesik Park, and Vladlen Koltun.
\newblock Fast global registration.
\newblock In {\em European Conference on Computer Vision (ECCV)}, 2016.

\end{thebibliography}
}

\end{document}